\documentclass{article}

\usepackage[preprint]{neurips_2024}

\usepackage[utf8]{inputenc} 
\usepackage[T1]{fontenc}    
\usepackage{hyperref}       
\usepackage{url}            
\usepackage{booktabs}       
\usepackage{amsfonts}       
\usepackage{nicefrac}       
\usepackage{microtype}      
\usepackage{xcolor}         
\usepackage{amssymb}            
\usepackage{mathtools}          
\usepackage{mathrsfs}           
\mathtoolsset{showonlyrefs}     
\usepackage{graphicx}           
\usepackage{subcaption}         
\usepackage[space]{grffile}     
\usepackage{url}                
\usepackage{todonotes}          
\usepackage{amsthm}             
\usepackage{wrapfig}            
\usepackage{bbm}                
\usepackage[noEnd]{algpseudocodex}      
\usepackage{algorithm}          
\usepackage{lipsum}
\usepackage{thmtools}           
\usepackage{thm-restate}        
\usepackage{bm}                 
\usepackage[export]{adjustbox}  

\DeclareMathOperator{\E}{\mathbb{E}}
\DeclareMathOperator{\KL}{\text{KL}}
\DeclareMathOperator{\diag}{\text{diag}}
\DeclareMathOperator*{\argmax}{\text{argmax}}

\title{Discovering Minimal Reinforcement Learning Environments}

\author{%
  Jarek Liesen \thanks{Correspondence at \texttt{jarek@bccn-berlin.de}}\\
  BCCN Berlin\\
  \And
  Chris Lu \\
  University of Oxford \\
  FLAIR \\
  \And
  Andrei Lupu \\ 
  University of Oxford \\
  FLAIR \\
  \AND
  Jakob~N. Foerster \\
  University of Oxford \\
  FLAIR \\
  \And
  Henning Sprekeler \\
  Technical University Berlin \\
  Science of Intelligence \\
  \And
  Robert~T. Lange \\
  Technical University Berlin \\
  Science of Intelligence \\
}

\begin{document}

\maketitle

\begin{abstract}
Reinforcement learning (RL) agents are commonly trained and evaluated in the same environment. In contrast, humans often train in a specialized environment before being evaluated, such as studying a book before taking an exam. The potential of such specialized training environments is still vastly underexplored, despite their capacity to dramatically speed up training.

The framework of \textit{synthetic environments} takes a first step in this direction by meta-learning neural network-based Markov decision processes (MDPs). The initial approach was limited to toy problems and produced environments that did not transfer to unseen RL algorithms. We extend this approach in three ways: 
Firstly, we modify the meta-learning algorithm to discover environments invariant towards hyperparameter configurations and learning algorithms. Secondly, by leveraging hardware parallelism and introducing a curriculum on an agent's evaluation episode horizon, we can achieve competitive results on several challenging continuous control problems. Thirdly, we surprisingly find that contextual bandits enable training RL agents that transfer well to their evaluation environment, even if it is a complex MDP. Hence, we set up our experiments to train \textit{synthetic contextual bandits}, which perform on par with synthetic MDPs, yield additional insights into the evaluation environment, and can speed up downstream applications.
\end{abstract}

\section{Introduction}
\label{sec:introduction}

\suppressfloats[t]
\begin{figure}
    \centering
    \includegraphics[width=\textwidth]{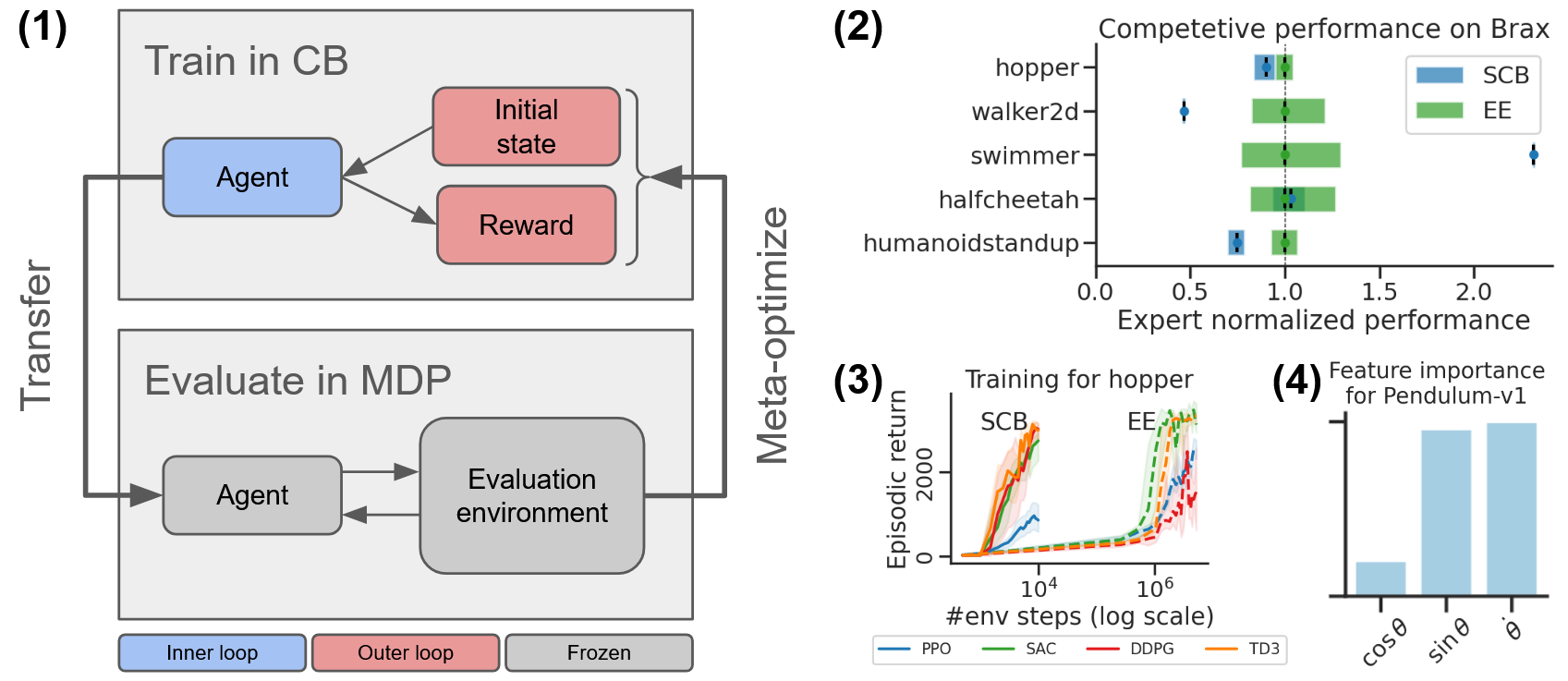}
    \caption{
    \textbf{(1)} Training process for synthetic contextual bandits. 
    Firstly, an agent is trained in an SCB (blue), observing only an initial state and reward in each episode. 
    After training, the agent is frozen and transferred to an evaluation environment. 
    The achieved episodic return is the training signal to update the SCB (red). 
    \textbf{(2)} Aggregated training results for challenging Brax environments. Training in the SCB yields policies that are competitive with EE experts, sometimes even outperforming them. The x-axis shows expert normalized performance as $(R - R_\text{random}) / (R_\text{expert} - R_\text{random})$, where $R_\text{random}$ and $R_\text{expert}$ are the episodic returns achieved in the EE by an expert and a random policy, respectively. For full training results, see appendix~\ref{tab:brax_retsults_table}.
    \textbf{(3)} Training curves for EE versus SCBs, saving orders of magnitude of environment steps. A complete visualization is given in figure~\ref{fig:training_one_row}.
    \textbf{(4)} Visualization of observation feature importances for Pendulum-v1. For details refer to section~\ref{sec:results-interpret}.
    }
    \label{fig:conceptual}
\end{figure}

Reinforcement learning (RL) agents are commonly trained and evaluated in precisely the same environment. It is well known that this approach has several significant disadvantages: RL agents are brittle to minor changes in the environment dynamics, hyperparameter choices, or even the concrete implementation of an algorithm \citep{henderson2018deep,engstrom2019implementation, cobbe2020leveraging,agarwal2021deep}.
Most recent research in RL has focused on improving RL algorithms to alleviate these challenges. But what about the RL environment or the underlying Markov decision process (MDP) itself? Unlike RL agents, professional athletes train under vastly different conditions than their final competition settings. For example, long-distance runners do not repeatedly run the target distance, but train shorter interval runs, progressively increase their pace, and occasionally mix in long runs. 
Such specialized training environments have the potential to significantly speed up RL pipelines:
They can be optimized to train agents rapidly, requiring several orders of magnitude fewer environment steps.  
Additionally, when such environment proxies are parameterized by neural networks, modern hardware accelerators enable rapid simulation of the environment.
Thus, they can be used in a myriad of applications including pretraining, neural architecture search, and downstream meta-learning.  
\\
Here, we scale the framework of \textit{synthetic environments} \cite[SEs,][]{ferreira2022learning} to explore the usefulness of synthetic training data for RL agent training.
SEs are RL environments parameterized by neural networks and optimized for transfer performance:
After training an agent in an SE, the agent achieves a high episodic return in a fixed evaluation environment (EE).
A visualization of the optimization algorithm is shown in figure~\ref{fig:conceptual} (1).
The initial approach parameterizes the SE using a single network to represent the transition, reward, and termination functions.  
Combined with the initial state distribution of the EE, the SE becomes a full MDP.
During our experiments, we found that extending the initial approach by additionally parameterizing the initial state distribution leads to synthetic MDPs that terminate most episodes after a single time step.  
We propose that this is not an artifact of meta-learning, but a discovered property that is beneficial when training agents.
Therefore, we purposefully constrain our environment parameterization towards synthetic contextual bandits (SCBs).
SCBs perform competitively with synthetic MDPs, but have several practical benefits including much smaller models and a high degree of interpretability.
We make the following contributions:

\begin{itemize}
    \item We give analytical and empirical evidence that it is possible to transform MDPs into CBs.
    In other words, we show that we can obtain a policy that performs well in an MDP by training it in a CB.
    We demonstrate that SCBs arise naturally when parameterizing fully synthetic MDPs (section~\ref{sec:discovering_scbs}).
    \item We show that our meta-learning algorithm discovers CBs invariant towards learning algorithms and hyperparameters, and even generalize towards out-of-distribution agents and training algorithms. Furthermore, we are the first to show that it is possible to discover synthetic proxies for challenging control environments ((2) and (3) in figure~\ref{fig:conceptual}, and section~\ref{sec:metalearning_scbs}).
    \item We demonstrate how the synthetic CBs can be analyzed to gain insights into their evaluation environment, including a measure of feature importance ((4) in figure~\ref{fig:conceptual}, and section~\ref{sec:results-interpret}).
    \item We show how SCBs can be integrated into downstream meta-learning applications, such as Learned Policy Optimization \citep{lu2022discovered}, speeding them up significantly (section~\ref{sec:dpo}).
    \item We implement several common RL algorithms in a way that supports GPU parallelism, which allows us to run experiments in a fraction of wall clock time compared to a multiprocessing-based approach.
    Additionally, we release the synthetic environments using the gymnax interface \citep{gymnax2022github}, allowing for a drop-in replacement of the evaluation environment. The respositories are available at \url{https://github.com/keraJLi/rejax} \citep{rejax} and \url{https://github.com/keraJLi/synthetic-gymnax}.
\end{itemize}

\section{Background}
\subsection{Contextual Bandits are a Special Case of Markov Decision Processes}
\label{sec:cbs_and_mdps}
The most widely used formalism for RL environments is the \textit{Markov decision process} (MDP), defined as a tuple $\langle S, A, P, R, \eta_0, d\rangle$. 
At the beginning of each episode, a state $s_0 \in S$ is sampled from the initial state distribution $s_0 \sim \eta_0$. 
At each succeeding time step $t$, the agent samples an action $a_t \in A$ from its policy $a \sim \pi(.|s_t)$. 
The environment then generates the next state as $s_{t+1} \sim P(.|s_t, a_t)$ and issues a reward $r_t \sim R(.|s_t, a_t)$.
As soon as the Boolean value of the termination function $d(s_t)$ indicates it, the episode is terminated.

The basic reinforcement learning problem is to find a policy $\pi^*$ that maximizes the expected return $\E_\pi[\sum_{t=0}^\infty\gamma^t r_t]$, where $0 < \gamma < 1$ is called discount factor.
Alternatively, this can be stated in terms of the Q-function as
\begin{align}
\pi^* &= \argmax_\pi \underset{\substack{s_0 \sim \eta_0 \\ a_0 \sim \pi(. | s_0)}}{\E}[Q^\pi(s_0, a_0)] \label{eq:pi_star}, \quad \text{where} \quad
Q^\pi(s, a) = \E_\pi\left[\sum_{t=0}^\infty \gamma^t r_t \middle| s_0 = s, a_0 = a\right]. 
\end{align}
Note that any optimal policy chooses actions greedily with respect to its Q-function, meaning $\pi^*(a | s) = 1 \text{ iff } a = \argmax_{\tilde a} Q^{\pi^*}(s, {\tilde a})$.

This work focuses on meta-learning a special case of MDPs, namely a \textit{contextual bandit} (CB). 
In a CB, the transition function $P$ is deterministic, and always points towards a state $s_d$ with $d(s_d) = \text{true}$.
From this constraint, it follows immediately that the Q-function of \textit{any policy} is equal to the expected immediate reward:
\begin{equation}
Q^\pi(s, a) = \E_\pi\left[\sum_{t=0}^\infty \gamma^t r_t \middle| s_0 = s, a_0 = a\right] = \E_\pi[\gamma^0 r_0 | s_0 = s, a_0 = a] = \E[R(. | s, a)]. \label{eq:cb_q_reward}
\end{equation}
Therefore in any contextual bandit, the optimal policy as defined in equation~\eqref{eq:pi_star} greedily maximizes the immediate reward, i.e. 
\[
\pi^*(a | s) = 1 \text{ iff } a = \argmax_{\tilde a} \E[R(.|s, {\tilde a})].
\]

\subsection{Meta-training Synthetic Environment and Parameterization}
\begin{algorithm}[h]
\caption{Bi-level optimization algorithm for meta-learning SEs}
\begin{algorithmic}
\Require number of generations $G$, population size $N$
\State Initialize population of SCBs via random neural network initialization
\For{$g = 1, \dots, G$} \Comment{Outer loop}
\For{$i = 1, \dots, N$} \Comment{Inner loop}
\State Train RL agent $A_i$ in $\text{SCB}_i$
\State Evaluate fitness of $A_i$ as episodic return in evaluation environment
\EndFor
\State Update population using meta-optimizer and performances of the agents $A_i$
\EndFor
\end{algorithmic}
\label{alg:meta_learning_ses}
\end{algorithm}

\citet{ferreira2022learning} introduce synthetic environments as RL environments parameterized by a neural network.
They parameterize parts of an MDP, namely the transition, reward, and termination function, computed as $s_{t+1}, r_t, d_t = f_\theta(s_t, a_t)$, where $f_\theta$ refers to the forward pass of a neural network with parameters $\theta$.
The resulting networks are then optimized using a bi-level optimization scheme consisting of two nested loops (see alg.~\ref{alg:meta_learning_ses}).
In the \textit{inner loop}, an RL agent is trained in a synthetic environment.
After training it is frozen, and its fitness, the episodic return in an evaluation environment, is calculated.
At each generation (iteration) of the \textit{outer loop}, the inner loop is executed on a population (batch) of SEs.
Afterward, the calculated fitness scores are used to generate the next population, such that the expected return is increased.
We use separable natural evolution strategies \citep[SNES,][see appendix~\ref{sec:meta_evo}]{wierstra2014natural} for outer loop optimization.


\section{Methods: Improving the Discovery of Synthetic Environments by Sampling Inner Loop Algorithms \& Introducing an Outer Loop Curriculum}
\label{sec:methods}
\textbf{Meta-learning for generalization by sampling algorithms.}
The meta-learned CBs should not be specific to certain RL algorithms. 
Instead, it should be possible for any RL algorithm to train a good policy in the SCB, using a wide range of hyperparameters (HPs).
To avoid overfitting to specific algorithms while meta-training, we extend the original optimization algorithm by sampling inner loop tasks, each of which is represented by a random algorithm/HP combination.
We use PPO, SAC, DQN and DDQN \citep{schulman2017proximal,christodoulou2019soft,Mnih2015,vanhasselt2015deep} for discrete, and PPO, SAC, DDPG and TD3 \citep{schulman2017proximal,haarnoja2018soft,lillicrap2015continuous,fujimoto2018addressing} for continuous action spaces.
HPs are sampled uniformly from a broad range of sensible values (see appendix~\ref{sec:hyper_inner}).

\textbf{Scaling to locomotion environments using an outer loop curriculum.}
Many continuous control problems in Brax \citep{freeman2021brax}, like hopper or walker2d, require learning balance and locomotion, and are truncated after 1000 steps.
When evaluating SCB-trained agents for the full 1000 steps, SCBs quickly converge to balancing without forward movement. 
To address this, we employ a curriculum on fitness evaluation rollout length: 
We start meta-training with short episodes and gradually increase their length, shifting focus towards locomotion early in meta-training.

\textbf{Leveraging automatic vectorization.} To efficiently parallelize the training of agent populations, we implement vectorizable versions of these algorithms in JAX \citep{jax2018github}.
This allows for hardware-parallel training using different values of hyperparameters that don't alter the memory layout or sequence of executed operations.
While this does not include model architecture or the number of training steps, we find that the diversity in training algorithms allows for sufficient generalization (section~\ref{sec:results-generalization}).
Additionally, we will publish the implementations as an open-source library, available at \url{https://github.com/keraJLi/rejax} \citep{rejax}.

\section{Results: Synthetic CBs are General \& Scalable MDP Proxies}
We first demonstrate that contextual bandits arise naturally from parameterizing fully synthetic MDPs (section~\ref{sec:discovering_scbs}).
Subsequently, we show that meta-learned synthetic contextual bandits generalize out-of-distribution and scale towards challenging control environments (section~\ref{sec:metalearning_scbs}).

\subsection{Contextual Bandits as a Discovered Property of Synthetic Environments}
\label{sec:discovering_scbs}
\begin{figure}
    \centering
    \hspace{0.015\textwidth}
    \includegraphics[width=0.38\textwidth]{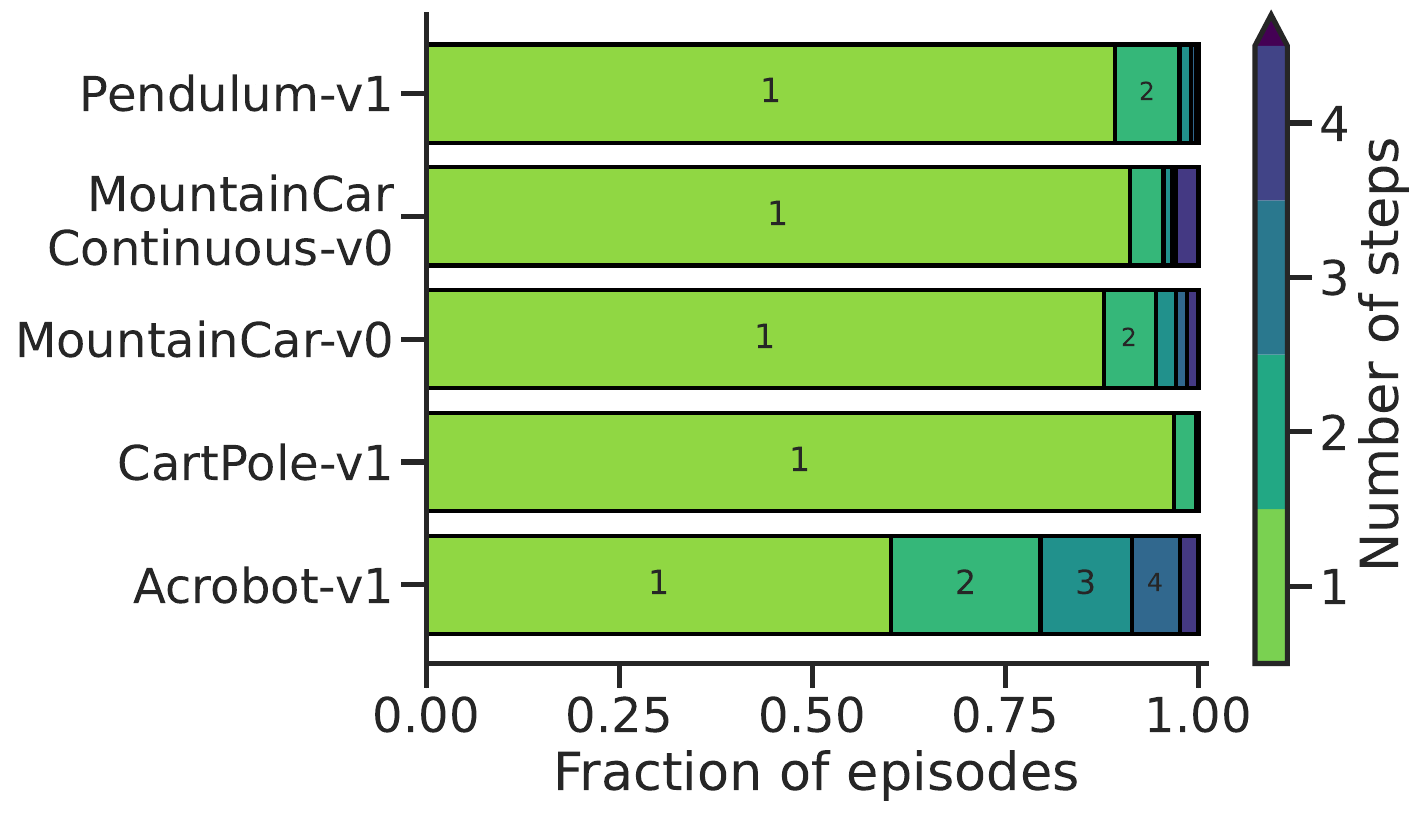}
    \hspace{0.06\textwidth}
    \includegraphics[width=0.42\textwidth]{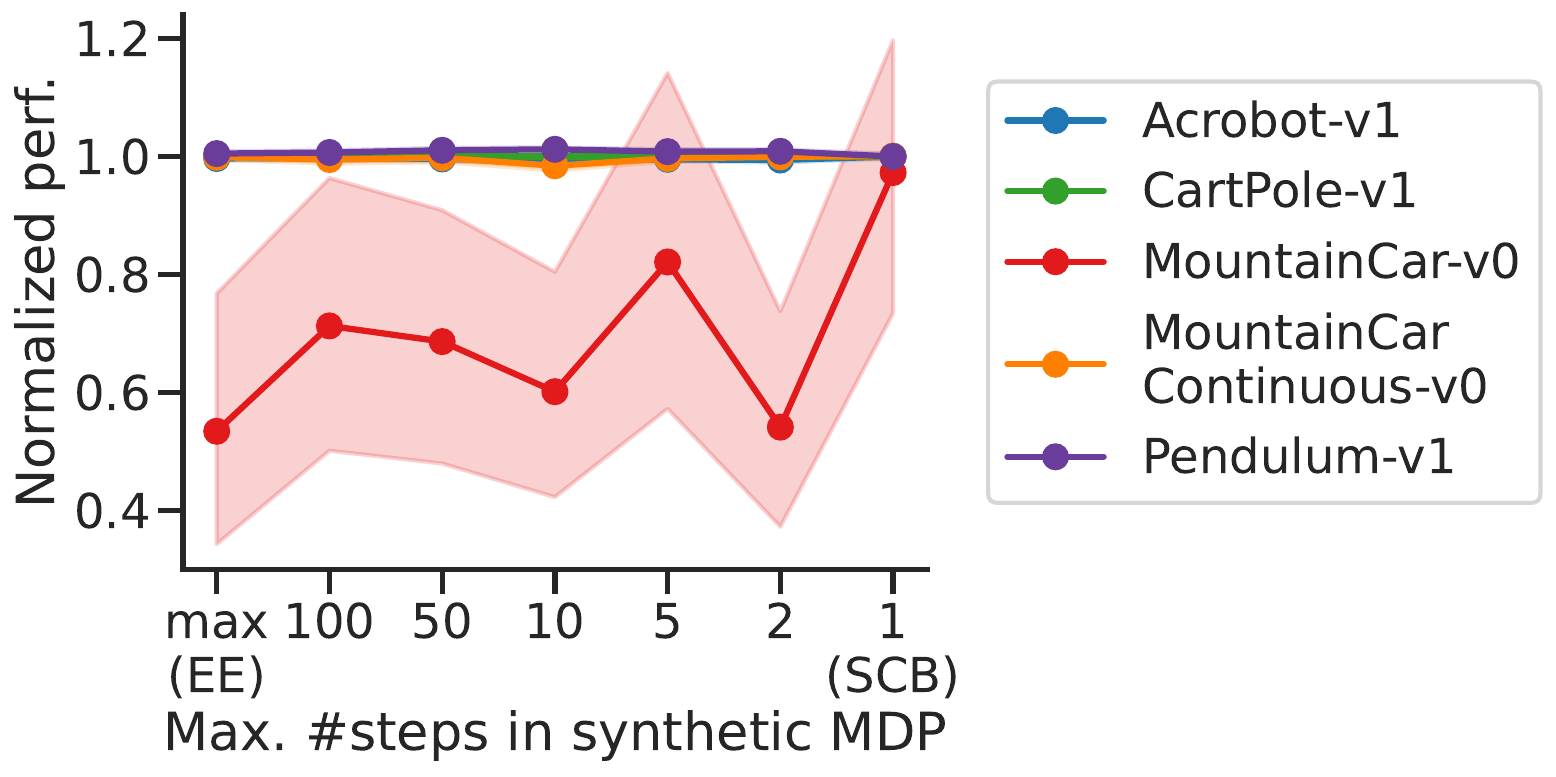}
    \hspace{0.015\textwidth}
    \caption{CBs are a discovered property of synthetic environments. \textbf{Left:} Fraction of episode lengths in a synthetic MDP. In most environments, more than 80\% of episodes are terminated after a single time step. Episodes were generated using 50 episodes of agents trained with each algorithm. \textbf{Right:} Normalized performance of synthetic MDPs with decreasing maximum episode length. ``max'' refers to the maximum episode of the evaluation environment. Shown are the IQM scores and 95\% confidence intervals of 20 training runs, aggregated over all algorithms (see section~\ref{sec:methods}). Performance is normalized as $(R - R_\text{SCB}) / (R - R_\text{random})$ for each algorithm individually, where $R_\text{SCB}$ is the return in the EE after training in the SCB, and $R_\text{random}$ is the return of a random policy.}
    \label{fig:episode_lengths}
\end{figure}

We begin our experiments by extending the setup of \citet{ferreira2022learning} with a parameterized initial state distribution of the form $s_0 = f_\phi(z)$, where $z \in \mathbb{R}^N$ is a latent variable sampled from a diagonal Gaussian distribution.
After training these fully synthetic MDPs, we found that they often terminate episodes after a single time step (see Figure~\ref{fig:episode_lengths}, left).
We interpret this as a \textit{discovered property} of SEs, and constrain our SEs to a single step, making them contextual bandits.
Surprisingly, we find that this has next to no negative impact on the performance of the synthetic environments, sometimes even being beneficial (see Figure~\ref{fig:episode_lengths}, right).
The usage of CBs has several practical advantages:
\begin{enumerate}
    \item The number of parameters is significantly lower.
    The transition function takes $\mathcal{O}(\dim(S)^2)$ parameters, while the reward function only takes $\mathcal{O}(\dim(S))$ when parameterizing with fully connected networks, where $S$ is the state space.
    \item It avoids instabilities related to the recurrent application of the transition function. 
    When allowing for long episodes, we consistently encountered overflowing state values (NaNs) in the early stages of meta-training (for example, in the ablations in appendix~\ref{sec:meta_evo_ablations}).
    \item It significantly simplifies the meta-learning problem for sparse reward environments. Parameterizing the initial state distribution is necessary to obtain a well-performing SCB for the MountainCar-v0 environment (appendix~\ref{sec:meta_evo_ablations}).
    We hypothesize that this is because critical states can be shown to the agent immediately, instead of having to be reached via multiple (meta-learned) transitions. 
    \item The synthetic reward function is well-interpretable since the episodic return is equal to the immediate reward (eq.~\eqref{eq:cb_q_reward}). Neural parameterization allows differentiation and the application of interpretability methods.
    We present two ways to interpret the CBs in section~\ref{sec:results-interpret}.
\end{enumerate}

Intuition suggests that being more general, MDPs have a richer class of solutions (i.e. optimal policies) compared to CBs.
Perhaps surprisingly, this is not the case:
\begin{restatable}{lemma}{cb}
\label{lemma}
Given any Markov decision process $M$, there exists a contextual bandit $B$, such that every policy $\pi^*$ that is optimal in $B$ is also optimal in $M$. For a proof see appendix~\ref{sec:lemma_proof}.
\end{restatable}
Theorem~\ref{lemma} makes no statement about the training efficiency in practice. 
Thus, we dedicate the rest of our experiments to empirically demonstrate that training in CBs is not only possible but beneficial.

\subsection{Meta-Learning Synthetic Contextual Bandits}
\label{sec:metalearning_scbs}
\begin{figure}
    \centering
    \includegraphics[width=0.9\textwidth]{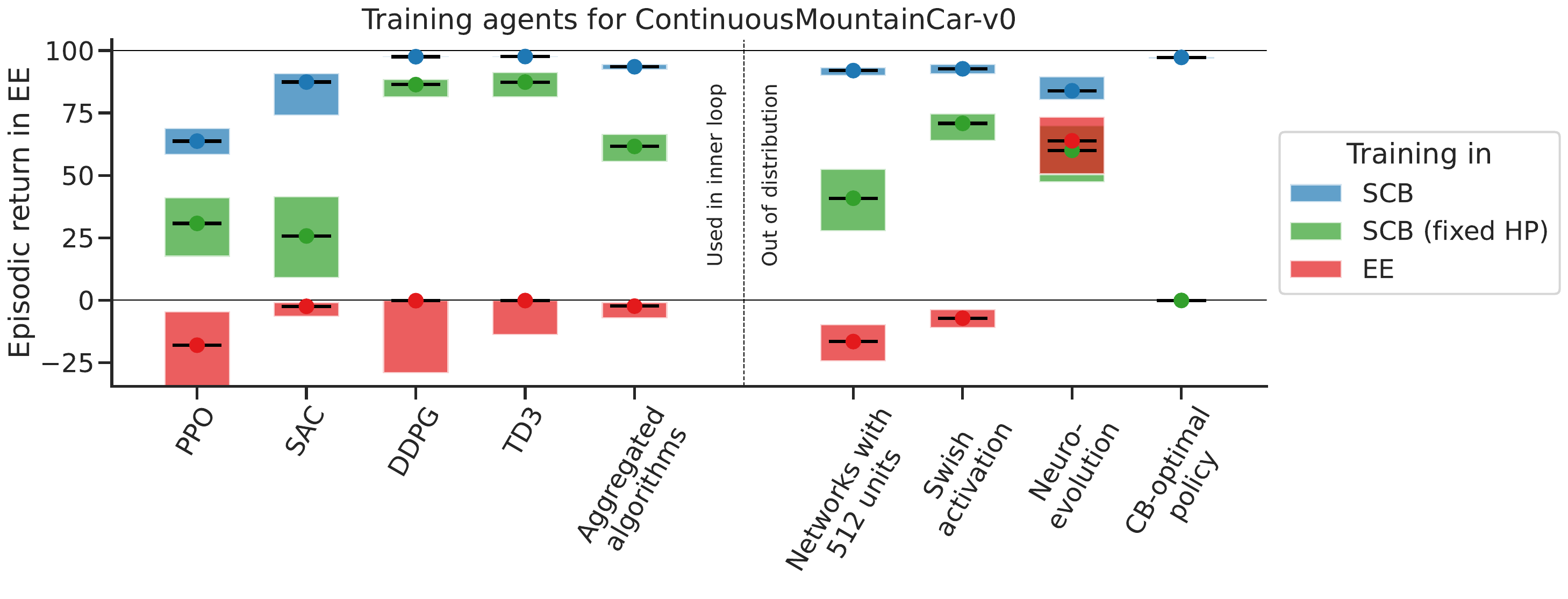}
    \caption{
    Meta-learned SCBs generalize across hyperparameters and towards out-of-distribution RL algorithms.
    Each column shows the return of a policy after training in either the SCB (blue) or the EE directly (red), using sampled hyperparameters and 10,000 environment steps.
    Additionally, we ablate the meta-training algorithm by using fixed hyperparameters in the inner loop. 
    When sampling hyperparameters during the evaluation, the ablated SCBs (green) perform worse.
    Additionally, the SCBs generalize towards agents not used in the inner loop (right of dashed line).
    For more details, refer to appenix~\ref{sec:generality_appendix}.
    }
    \label{fig:generalization}
\end{figure}

\textbf{SCB generalization.}
\label{sec:results-generalization}
Figure~\ref{fig:generalization} demonstrates the generality of a meta-learned SCB for ContinuousMountainCar-v0.
It shows the performance of inner loop agents with random HPs, as well as the performance of several agents that were out-of-distribution, after training in the SCB and evaluation environment directly.
SCBs are universally much more robust to hyperparameter changes than their respective evaluation environment.
This is even the case when being meta-learned using fixed hyperparameters in the inner loop, but the robustness can be increased further by sampling them.
Notably, SCBs enable consistently training PPO agents for MountainCar-v0, which is not the case in the evaluation environment directly, even when using tuned HPs\footnote{MountainCar-v0 has a sparse reward that is unlikely to be reached by random exploration. Thus, even standard reference implementations of PPO struggle to reach the goal. For example, this is the case for CleanRL \citep{huang2022cleanrl}, see \url{https://docs.cleanrl.dev/rl-algorithms/ppo/\#experiment-results} and appendix~\ref{sec:generality_appendix}}.
Due to the technical constraints imposed by using JAX's vectorization, SCBs were trained using both a fixed set of gradient-based algorithms and a common fixed network architecture for each RL agent.
Still, the meta-learned SCBs generalize \textit{out-of-distribution}.
Replacing an RL agent's activation function or network architecture with one that was not used during meta-training does not hurt its performance.
Additionally, optimizing a policy network using SNES \citep{wierstra2014natural}, an evolution strategy instead of a gradient-based algorithm, works well across SCBs.
Finally, we show that the SCB-optimal policy performs well in the evaluation environment, meaning that SCBs should enable training capable agents using any RL algorithm that finds this policy.

\textbf{Baselines and Ablations.}
Since CBs lack the temporal dynamics of an MDP, training in a CB can be thought of as an (online) supervised learning problem. Instead of solving the temporal credit assignment problem, the agent simply has to predict which action maximizes the immediate reward. We, therefore, investigate how the discovered reward function compares with several baselines.
First, we compare to online supervised learning of an expert policy, implemented by interacting with the evaluation environment and taking steps to minimize $\KL[\pi || \pi_\text{expert}]$ on batches of states.
We refer to this setup as online behavioral cloning and interpret it as a replacement for the reward function of the SCB.
Alternatively, we can replace the reward function with one that was constructed using an expert Q-function.
We can construct it such that theorem~\ref{lemma} holds, thereby theoretically obtaining expert policies from training in the SCB.
Finally, we can replace the synthetic initial state distribution with an expert state distribution, simulating the interaction of the evaluation environment and expert agent in the background. 
We find that the synthetic initial state distribution is strictly required for successful RL training, while the training speed of SCB training and online behavioral cloning is comparable (appendix~\ref{sec:replacing_components}).
In appendix~\ref{sec:meta_evo_ablations} we additionally perform several ablations to the most important design choices of our meta-training algorithm:
A parameterized initial state distribution is needed for sparse-reward environments, our method performs best with a continuous latent distribution for the initial state, and meta-evolution is robust to the choice of meta-curriculum.
\begin{figure}
    \centering
    \includegraphics[width=0.7\textwidth,valign=t]{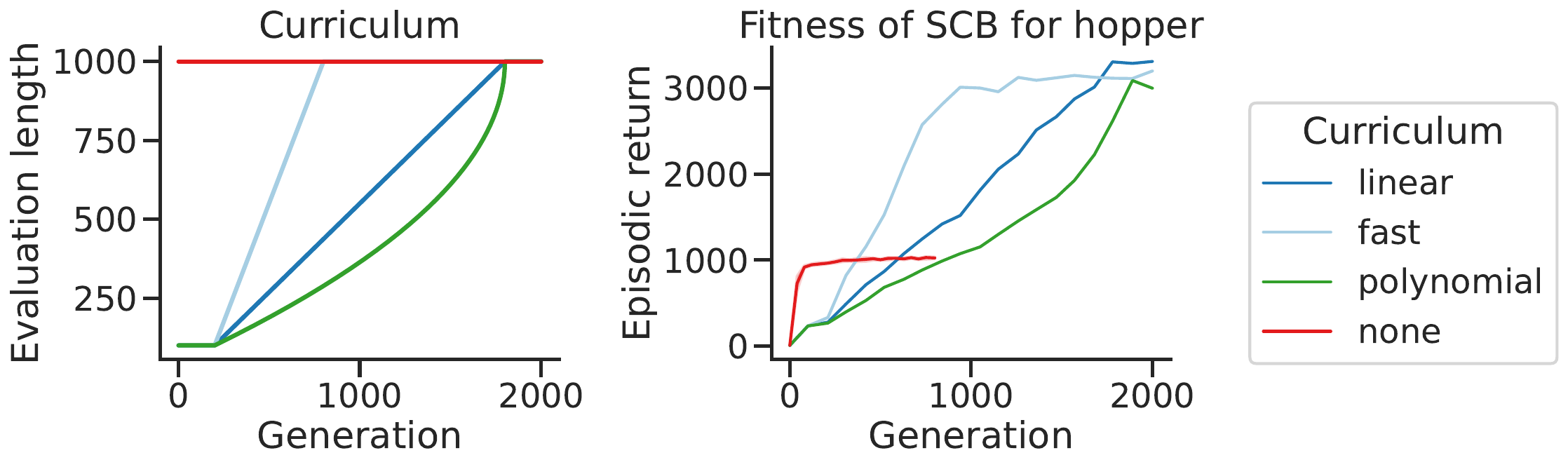}
    \caption{
    Curriculum and training speed for Brax environments. 
    \textbf{Left:} Visualization of different curricula. 
    \textbf{Right:} Influence of curricula on the meta-training progress for the hopper environment. For no curriculum, we show the IQM and 95\% CIs over 9 seeds. One seed is shown per curriculum.
    }
    \label{fig:brax}
\end{figure}

\begin{figure}
    \centering
    \includegraphics[width=\textwidth]{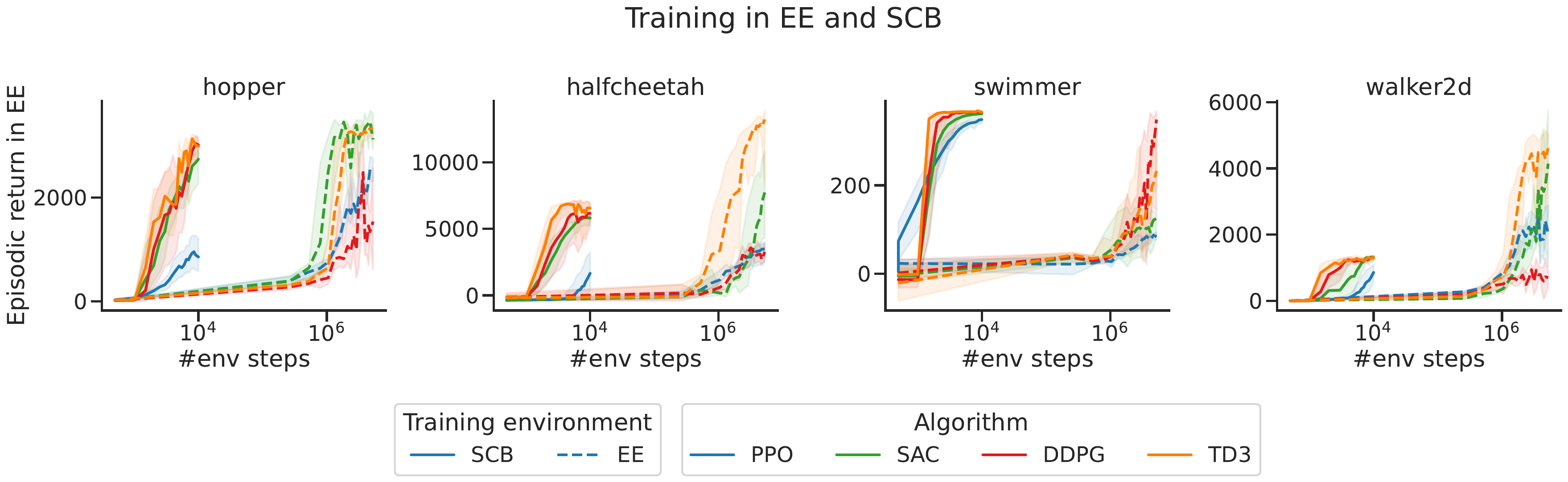}
    \caption{Training curves for different algorithms on Brax environments. Training in an SCB is roughly two orders of magnitude faster.
    We show the IQM performance with 95\% bootstrapped confidence intervals.
    Refer to appendix~\ref{sec:training_speed_comparison_brax} for complete training curves.}
    \label{fig:training_one_row}
\end{figure}

\textbf{Scaling to complex control environments.}
To scale to control environments requiring locomotion, we apply a curriculum to the meta-learning algorithm.
It gradually increases the number of time steps for which we evaluate a trained inner loop agent in the evaluation environment.
Different curricula and corresponding meta-learning curves are shown in figure~\ref{fig:brax}.
While our experiments suggest that meta-learning is robust towards the choice of curriculum, training without a curriculum leads to quick convergence to a local optimum.
Overall, this method allows us to successfully meta-learn SCBs for several complex control environments, as shown in figure~\ref{fig:conceptual} (2).
Notably, agents only take 10,000 steps to learn these tasks in the SCB, whereas training in the evaluation environments typically takes millions of time steps.
Training in the EE directly is roughly two orders of magnitude slower, as shown in figure~\ref{fig:training_one_row}.
Additionally, achieving good returns on Brax environments typically needs extensive hyperparameter tuning and additional hacks such as observation normalization, which is unnecessary when training on the SCB.

\section{Interpretability of Synthetic Contextual Bandits}
\label{sec:results-interpret}
\begin{figure}
\centering
\includegraphics[width=0.95\textwidth]{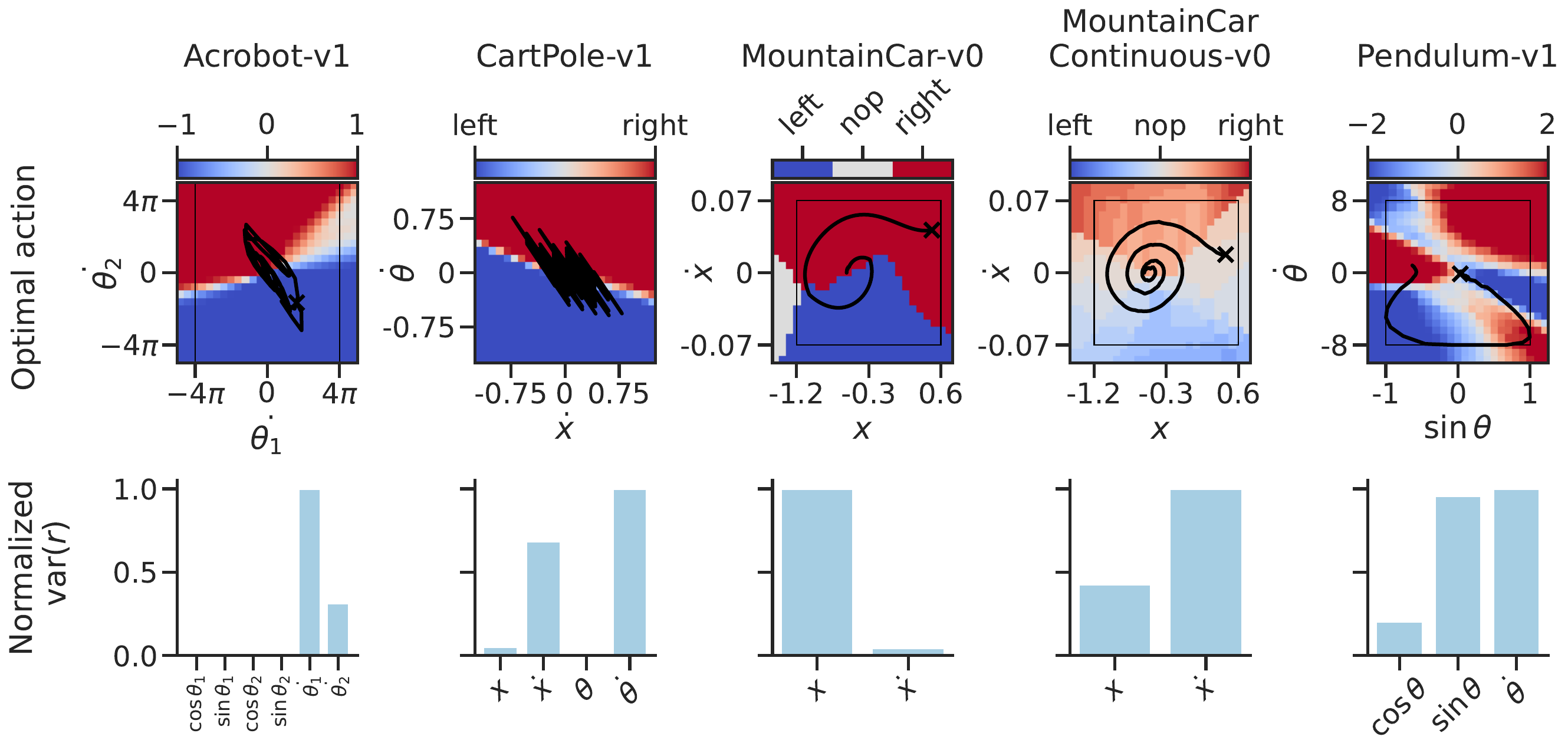}
\caption{
SCBs provide interpretable insights into their evaluation environment. 
\textbf{Top}. Optimal actions given the differentiable synthetic reward function for different states and 5 environments. 
Black box: observation space of the evaluation environment. 
Black line: representative trajectory in the real environment. 
Black x-marker: episode end. 
\textbf{Bottom}. Normalized reward variance when varying observation parts. 
Mean value over all observations in the space visualized in the top row.
}
\label{fig:interpretablility}
\end{figure}

In contextual bandits, the reward received is equal to the return, the state-, and the state-action value function (see eq.~\eqref{eq:cb_q_reward}). 
This enables new ways to analyze the environment, such as easily finding the optimal action in each state via gradient descent or a simple grid search.
We visualize the optimal actions in the top row of figure~\ref{fig:interpretablility}.
The resulting visualizations yield insights into the way that the synthetic environment trains an agent to perform a task:
For example, the SCB for MountainCar-v0 never induces nops, since the return is highest if terminating early, while the optimal action in the MountainCarContinuous-v0 SCB is often close to nop since it includes a control cost instead of a constant negative reward.
Additionally, we can directly investigate the relationship between the observation and the return.
We do so by fixing observation and action, and observing the variance in the reward when varying a single entry of the observation.
The results are visualized in the bottom row of figure~\ref{fig:interpretablility}.
We find that the reward is almost invariant to some parts of the observations.
For example, varying the values of the angle in Acrobot-v1 has very little impact on the reward compared to the angular velocities.
Similar findings hold for the position and angle in CartPole-v1.
Thereby we rediscover the results of \citet{vischer2021lottery} and \citet{lu2023adversarial}.
They found the same invariances in the context of the lottery ticket hypothesis and adversarial attacks respectively, where these input channels were pruned or used to manipulate learning dynamics.

\section{Downstream Application: Learned Policy Optimization}
\label{sec:dpo}

\begin{figure}
    \centering
    \includegraphics[width=\textwidth]{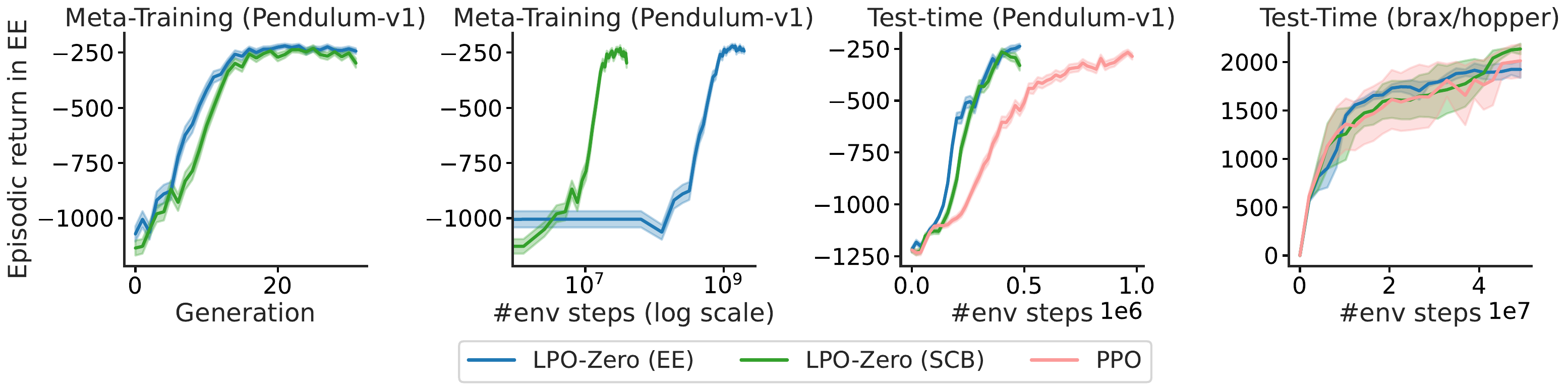}
    \caption{Meta-learning an objective function using an SCB. \textbf{From left to right:} \textbf{1.} Meta-training using SCB and EE achieves similar performance on Pendulum-v1. \textbf{2.} Meta-training on an SCB takes fewer environment steps since every episode has only one step. \textbf{3.} Training with the meta-learned objective is faster than using the original PPO objective. \textbf{4.} The meta-learned objectives generalize to an unseen, much more complex, environment (Hopper).}
    \label{fig:dpo_synthenv}
\end{figure}

SCBs offer a significant training speedup both in terms of environment steps and simulation time.
This offers several possibilities for downstream applications, such as in evolutionary meta-learning.
As an example, we demonstrate their effectiveness in Learned Policy Optimization \citep[LPO,][]{lu2022discovered}.
In LPO, the surrogate objective of PPO is replaced by a neural network\footnote{The neural network is parameterized to fulfill the conditions of a drift functional to uphold theoretical convergence guarantees, for more details see \citep{lu2022discovered,kuba2022mirror}.}.
This neural network is meta-trained to maximize the final return when using it as an objective.

We meta-learn an objective function both on the Pendulum-v1 environment, as well as an SCB that has Pendulum-v1 as its corresponding EE.
In both cases, we obtain well-performing objective functions (figure~\ref{fig:dpo_synthenv}, 1.).
However, meta-training on the SCB takes two orders of magnitude fewer environment steps (figure~\ref{fig:dpo_synthenv}, 2.).
This is because to obtain fitness scores for the SCB, only a single environment step has to be simulated, while for Pendulum-v1 a whole episode of 200 steps is necessary.
We then use the meta-learned surrogate objectives for training an agent on Pendulum-v1 and compare them to PPO (figure~\ref{fig:dpo_synthenv}, 3.).
We find that both objectives outperform the original objective of PPO in terms of training speed and final performance.
To probe the generalization to new environments, we additionally evaluate the meta-learned objectives on hopper (figure~\ref{fig:dpo_synthenv}, 4.).
The results show that the meta-learned objectives are fully comparable to PPO.

In this application, replacing the evaluation environment with its corresponding SCB had no negative effect on the final performance, but allowed for a significant speedup in meta-training time.
The ability to not only train an RL agent but to successfully apply a meta-learning algorithm on the SCB further underscores its generality.

\section{Related Work}
\textbf{Training Reinforcement Learning Agents with Synthetic Data}. 
Various methods for training machine learning models from synthetically generated data have been proposed. 
For example, this includes dataset distillation for supervised training \citep{wang2018dataset} or synthetic experience replay and behavior distillation for RL \citep{lu2023synthetic,lupu2024behaviour}. 
Applications for training with synthetic data include data augmentation and cheap data generation, which is especially important when requiring large amounts of data, such as in RL.
Most closely related to our work is the approach outlined by \citet{ferreira2022learning} which learns the reward- and state transition function while using the initial state distribution of the original environment. 
They report limited transfer to agents not used in the inner loop and do not scale their approach to continuous control environments.

\textbf{Extensions to \citet{ferreira2022learning}}
We overcome the limited scalability and transfer reported by \citet{ferreira2022learning} by extending their work in the following ways:
\begin{itemize}
    \item We limit the episode length in the synthetic MDP to one, turning it into a simple CB.
    \item We meta-learn the initial state distribution instead of sampling from the EE.
    \item We sample inner-loop algorithms during meta-training. This allows for broad generalization, even out-of-distribution, as shown in figure~\ref{fig:generalization}.
    \item We introduce a curriculum on the evaluation episode length for control environments. This enables training SCBs for complex control environments of the Brax suite, as shown in figure~\ref{fig:brax}.
    \item We leverage hardware acceleration by developing an efficient RL algorithm implementation. This allows us to run much larger experiments, using larger population sizes (256 vs. 16), more evaluation seeds (64 vs. 10), and more generations (2000 vs. 200).
\end{itemize}

\textbf{Discovering Algorithm Components via Evolutionary Meta-Learning}. Recently, the general combination of evolutionary optimization and neural network-based algorithm families has been used to discover various powerful algorithms. This includes the meta-discovery of gradient-based \citep{metz2022velo} and gradient-free \citep{les2022, lga_2023} optimization algorithms, policy optimization objective functions \citep{lu2022discovered, jackson2024discovering}, or reward functions \citep{faust2019evolving}. Furthermore, these synthetic artifacts can often be reverse-engineered to generate human-interpretable components. Here, we use the same paradigm to transform real environment simulators into SCBs.


\textbf{Hardware Accelerated Reinforcement Learning Environments}. Commonly, RL environments have been bound to CPUs and constrained by limited parallelism. Recently, there has been a paradigm change with RL simulators being accelerated by accelerator parallelism. These efforts include Brax \citep{freeman2021brax}, Gymnax \citep{gymnax2022github}, Jumanji \citep{bonnet2023jumanji}, Pgx \citep{koyamada2023pgx}, or NVIDIA Isaac Gym \citep{makoviychuk2021isaac}. Still, most of them require the translation of the original step transition logic into hardware-specific coding frameworks (e.g. JAX \citep{jax2018github}). Here, we provide a means to automatically yield hardware-accelerated neural-network-based environment proxies for training RL agents that generalize to potentially non-accelerated environments.

\section{Discussion}
\label{sec:discussion}
\textbf{Summary}. We have demonstrated that it is possible to transform Markov decision processes into contextual bandits, a much simpler class of reinforcement learning environments. 
The meta-learned SCBs are capable of training RL agents that perform competitively in evaluation environments. 
As shown by our extensive studies, they exhibit a high degree of generality and even generalize to RL agents out-of-distribution. 
To enable successful meta-learning, we introduced improvements over the previous discovery process by \citet{ferreira2022learning}, including the sampling of inner loop algorithms, a curriculum on the evaluation episode length, and an efficient implementation. 
The SCBs yield insights into the relevance of individual observation entries, are easy to interpret, and can be used to speed up downstream applications.

\textbf{Limitations}. 
Our goal for this project was to demonstrate the possibility of transforming Markov decision processes into contextual bandits, enabling fast training.
Still, optimizing an SCB using black-box meta-learning is far more computationally expensive than training agents in the evaluation environment directly.
Current limitations of black-box meta-learning also extend to this work, limiting the number of trainable parameters in practice.
To recoup the high initial cost, the SCB has to be used in downstream applications, like Learned Policy Optimization \citep{lu2022discovered}.
While a curriculum was necessary to discover SCBs for Brax environments, other hacks might be necessary for different classes of more complex tasks, all of which must be engineered.

\textbf{Future Work}. 
Going forward we are interested in the discovery of synthetic simulators capable of promoting a truly open-ended learning process. 
Furthermore, we have focused on control environments with proprioceptive symbolic observation dimensions so far. 
A natural extension of our work is to pixel- and vision-based environments leveraging transposed convolutional architectures for the initial state distribution.

\textbf{Societal impact}. We find that neural networks are capable of representing various RL simulators in a compressed fashion. In principle, large models can therefore be capable of distilling data distributions and world models useful for self-training. Given that these systems are ultimately black-box, practitioners need to be careful when deploying them in real-world applications.

\begin{ack}
Andrei Lupu was partially funded by a Fonds de recherche du Québec doctoral training scholarship.
Henning Sprekeler and Robert T. Lange are funded by the Deutsche Forschungsgemeinschaft (DFG, German Research Foundation) under Germany’s Excellence Strategy - EXC 2002/1 ``Science of Intelligence'' - project number 390523135. 
\end{ack}

\bibliographystyle{plainnat}
\bibliography{main}
\clearpage


\appendix
%
%

\section{Solving Markov Decision Processes by Training in Contextual Bandits}
\label{sec:lemma_proof}
\cb*
\vspace{-1em}
\begin{proof}
Let $\pi_M^*$ be an optimal policy in $M$ with value function $Q^*_M(s, a)$.
We construct $B$ by setting $S_B = S_M$ and $A_B = A_M$, where $S$ and $A$ are state and action spaces.
Furthermore, we set $R_B(s, a) = Q^*_M(s, a)$ and $\rho_{0,B} = \rho_M(\pi_M^*)$, where $\rho_M(\pi_M^*)$ is the state distribution of policy $\pi_M^*$ interacting with environment $M$.

We now show that any policy $\pi_B^*$ that is optimal in $B$ is also optimal in $M$, by noting that
\begin{enumerate}
    \item $\pi^*(a|s) = \argmax_a Q^*(s, a)$ for any policy,
    \item $\pi^*_B(a|s) = \argmax_a R_B(s, a)$ because $Q_B^*(s, a) = R_B(s, a)$ (see equation~\eqref{eq:cb_q_reward}), and
    \item $R_B(s, a) = Q^*_M(s, a)$ by construction.
\end{enumerate}

Therefore, for all states $s$ that $\pi_B^*$ visits while interacting with $M$,
\[
\pi^*_B(a | s) \overset{2.}{=} \argmax_a R_B(s, a) \overset{3.}{=} \argmax_a Q^*_M(s, a) \overset{1.}{=} \pi^*_M(a | s). \qedhere
\]
\end{proof}

There are several different constructions that can be used in the proof, as $R_B$ only has to be equal to $Q^*_M$ under $\argmax$ (see equality 3.).
For example, one can use $R_B(s, a) = - ||a - \argmax_{\tilde a} Q^*_M(s, {\tilde a})||$ for continuous, and $R_B = \mathbbm{1}[a = \argmax_{\tilde a} Q^*_M(s, {\tilde a})]$ for discrete environments.
Additionally, one can use any initial state distribution that has non-zero probability for all states visited by $\pi_M^*$.
We compare to these baselines in appendix~\ref{sec:replacing_components}.

\section{Meta-Optimization via SNES}
\label{sec:meta_evo}
In our experiments, we use a variant of evolution strategies \citep[ES,][]{wierstra2014natural} for meta-optimization.
ES are algorithms for black-box optimization inspired by the process of biological evolution. 
They aim to maximize a fitness score by sampling a population of parameters and mutating those with high fitness to create a population for the next iteration.
Formally, every ES consists of a parameterized search distribution $\pi(\bm{z}|\theta)$ and a fitness evaluation function $f(\bm{z})$.
Here $\bm{z}$ is a vector of parameters whose fitness to maximize, and $\theta$ are the parameters of the search distribution.
The goal is to maximize the expected fitness under the search distribution 
\[
J(\theta) = \mathbb{E}_\theta[f(\bm{z})] = \int f(\bm{z}) \pi(\bm{z}|\theta) d\bm{z}.
\]

The key design choices of ES are the parameterization of the search distribution and the update of its parameters.
We use Separable Natural ES \citep[SNES,][]{wierstra2014natural}, which updates the search distribution by approximating the natural gradient of $J(\theta)$ with respect to $\theta$.
The search distribution of SNES is a diagonal Gaussian $\pi(\bm{z}|\theta) = \mathcal{N}(\bm{z}|\bm{\mu}, \bm{\sigma})$ where $\bm{\sigma} = \diag(\sigma_1, \dots, \sigma_M)$ and $M$ is the number of parameters.
A generic version of SNES is shown in algorithm~\ref{alg:snes}.
Commonly, SNES additionally applies fitness shaping to become invariant to monotonic transformations of the fitness vector. We use the implementation provided by Evosax \citep{evosax2022github}.
For more details, see \citet{wierstra2014natural}.

\begin{algorithm}
\begin{algorithmic}
\Require Population size N
\Require Fitness function $f(\bm{z})$
\State Initialize search distribution mean $\bm{\mu}$ and diagonal covariance matrix $\bm{\sigma}$
\While{not converged}
\State 1. Sample population $\bm{z}_i \sim \mathcal{N}(-|\bm{\mu}, \bm{\sigma})$ for $i = 1, \dots, N$
\State 2. Calculate fitness $f(\bm{z}_i)$ and log-gradients $\log \nabla_\theta \pi(\bm{z}|\theta)$
\State 3. Estimate $\nabla_{\bm{\mu}} J(\theta) \approx \sum_{i=1}^N f(\bm{z}_i) \bm{s_i}$ and $\nabla_{\bm{\sigma}} J(\theta) \approx \sum_{i=1}^N f(\bm{z}_i) (\bm{s}_i^2 - 1)$, where $\bm{s}_i = \frac{\bm{z}_i - \bm{\mu}}{\bm{\sigma}}$
\State \hphantom{\widthof{3. }}Update $\bm{\mu} \gets \bm{\mu} + \eta_{\bm{\mu}} \bm{\sigma} \nabla_{\bm{\mu}}J$ and $\bm{\sigma} \gets \bm{\sigma} \exp (\frac{\eta_{\bm{\sigma}}}{2} \nabla_{\bm{\sigma}}J)$
\EndWhile
\end{algorithmic}
\caption{Separable NES}\label{alg:snes}
\end{algorithm}

\pagebreak
\section{Generality of Synthetic Contextual Bandits}
\label{sec:generality_appendix}
Here, we present results on the generality of SCBs for other classic control environments, analogous to ContinuousMountainCar-v0 figure~\ref{fig:generalization}.
We train each agent on the x-axis for 20 independent runs and show the IQM and 95\% confidence intervals of the episodic return achieved in the EE.
Each agent's performance is evaluated using the mean return of 50 episode rollouts.
For the inner loop algorithms, we sample hyperparameters from the distribution used in the inner loop.
We compare training using our standard SCB (blue), an SCB that was meta-learned using fixed hyperparameters in the inner loop (green), and the EE (red).
``Aggregated algorithms'' refers to the episodic return in the EE aggregated over all inner loop algorithms (PPO, SAC, DDPG, TD3 for continuous, and PPO, SAC, DQN for discrete action spaces).
To test for generalization out of distribution, we train several different agents that were not included in the inner loop.
These are:
\begin{description}
\item[Networks with 512 units] Agent networks with a different architecture (one hidden layer with 512 units instead of two hidden layers with 64 units). Aggregated across all inner loop algorithms.
\item[Swish activation] Agent networks with a different activation function (swish instead of tanh). Aggregated across all inner loop algorithms.
\item[Neuroevolution] Training a policy network using a gradient-free evolution strategy instead of a gradient-based RL algorithm. Since neuroevolution was not used in the inner loop, we have not defined a hyperparameter distribution to sample from, using default hyperparameters instead.
\item[CB-optimal policy] We evaluate the optimal policy in the CB, which we find by gradient-based maximization of $\argmax_a R_B(s, a)$ for each state $s$ for continuous action spaces, and by explicitly calculating $\argmax_a R_B(s, a)$, trying out all actions, for discrete action spaces. Since this policy is not trained, there are no training hyperparameters that can be sampled.
\end{description}

\begin{figure}[H]
    \centering
    \includegraphics[width=0.8\textwidth]{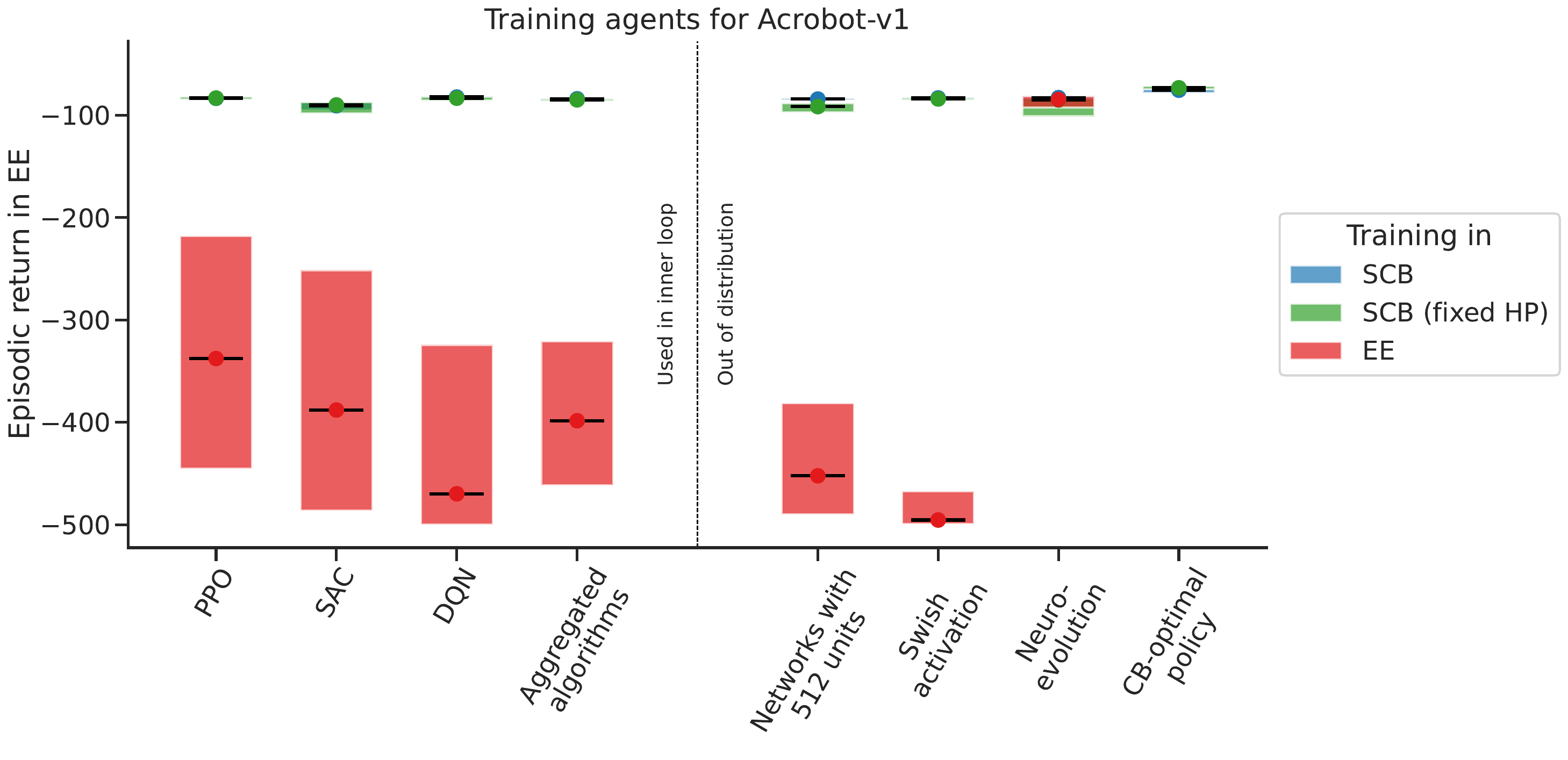}
    \caption{Generality of an SCB for Acrobot-v1.}
\end{figure}
\begin{figure}[H]
    \centering
    \includegraphics[width=0.8\textwidth]{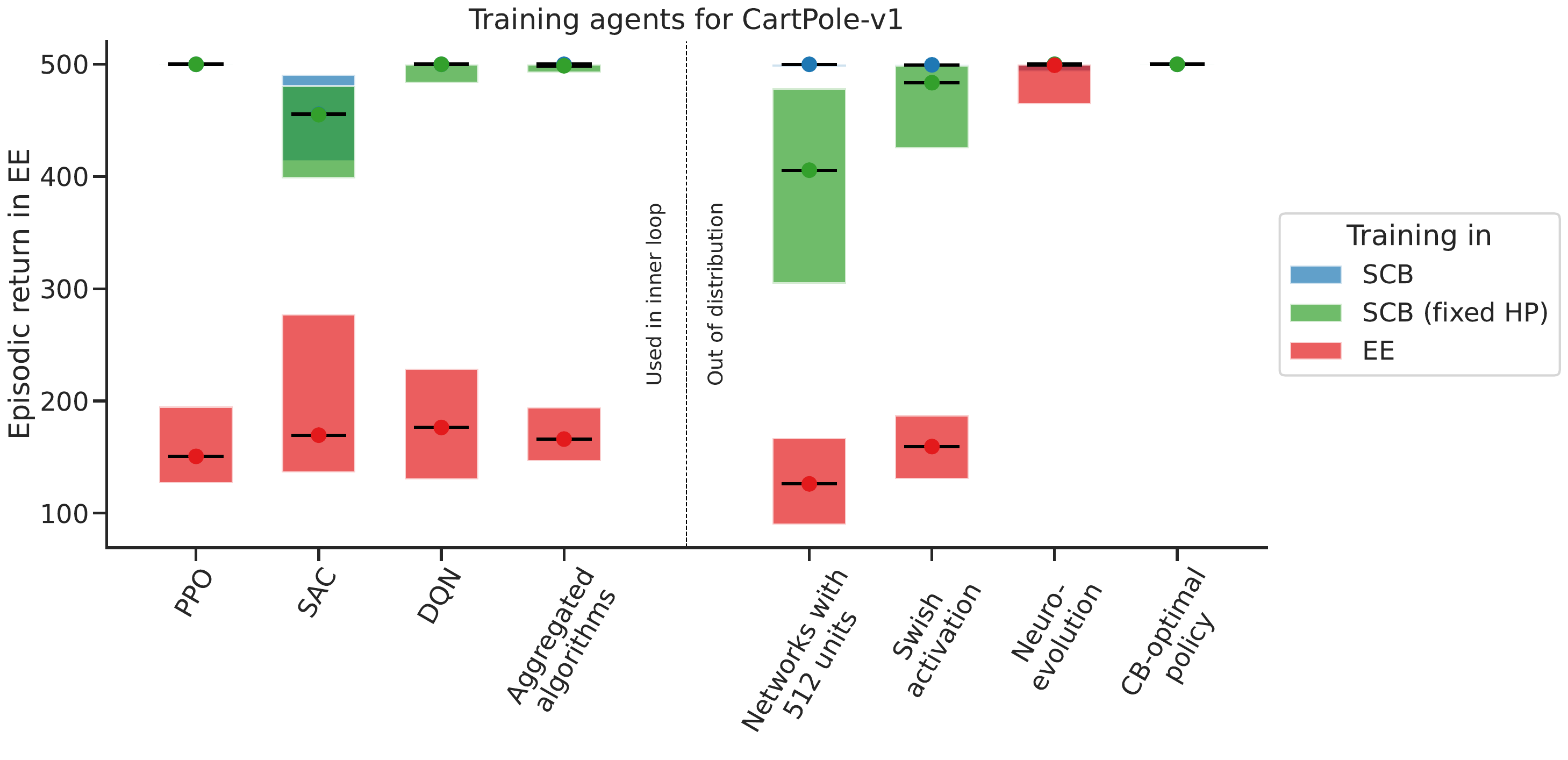}
    \caption{Generality of an SCB for CartPole-v1. The environment is considered solved if the return equals 500.}
\end{figure}
\begin{figure}[H]
    \centering
    \includegraphics[width=0.8\textwidth]{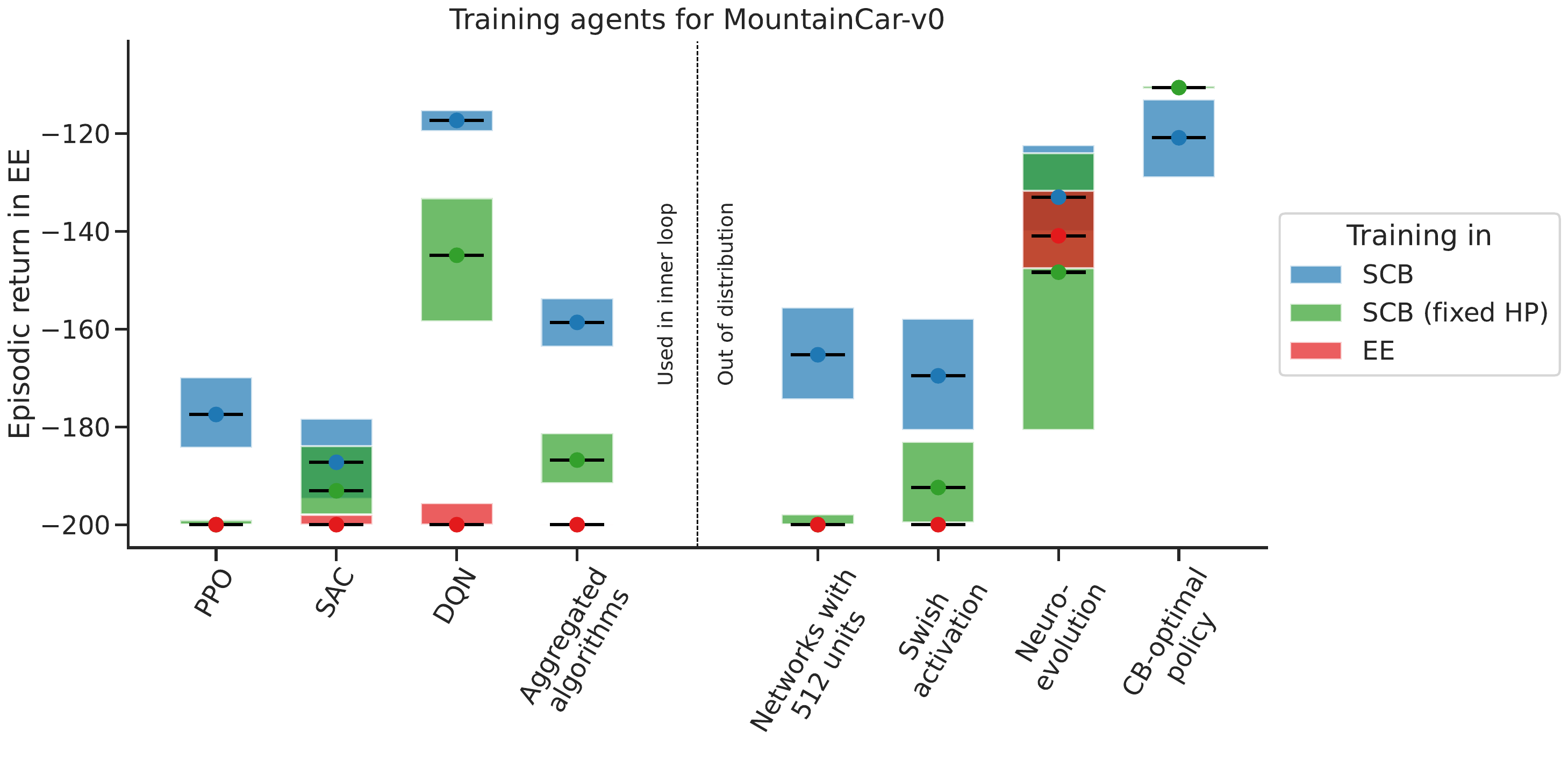}
    \caption{Generality of an SCB for MountainCar-v0. The return being $>200$ indicates that the goal was reached.}
\end{figure}
\begin{figure}[H]
    \centering
    \includegraphics[width=0.8\textwidth]{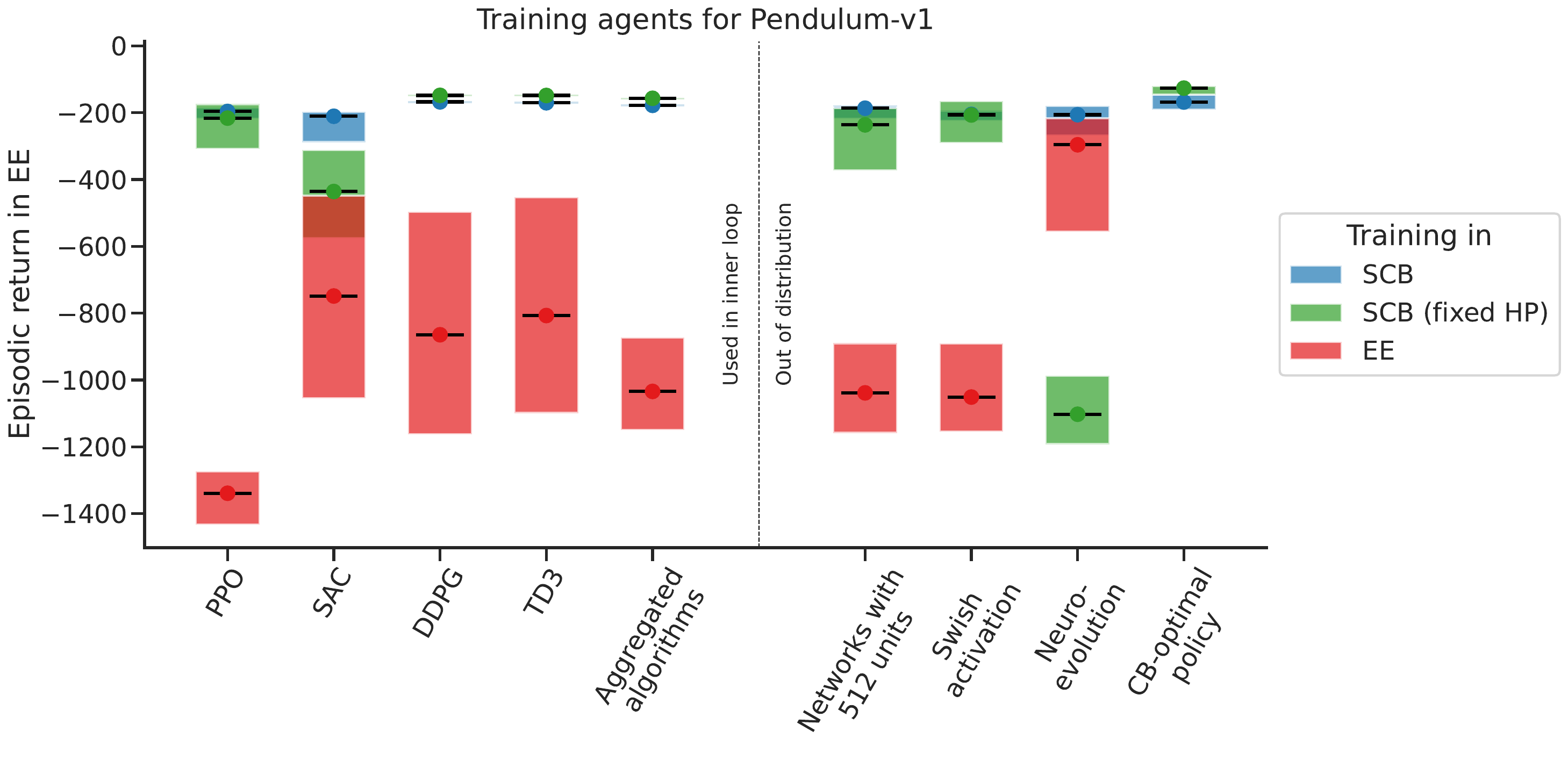}
    \caption{Generalty of an SCB for Pendulum-v1.}
\end{figure}

\section{Ablations}
\subsection{Replacing Components of an SCB}
\label{sec:replacing_components}
\begin{figure}[H]
    \centering
    \includegraphics[width=0.9\textwidth]{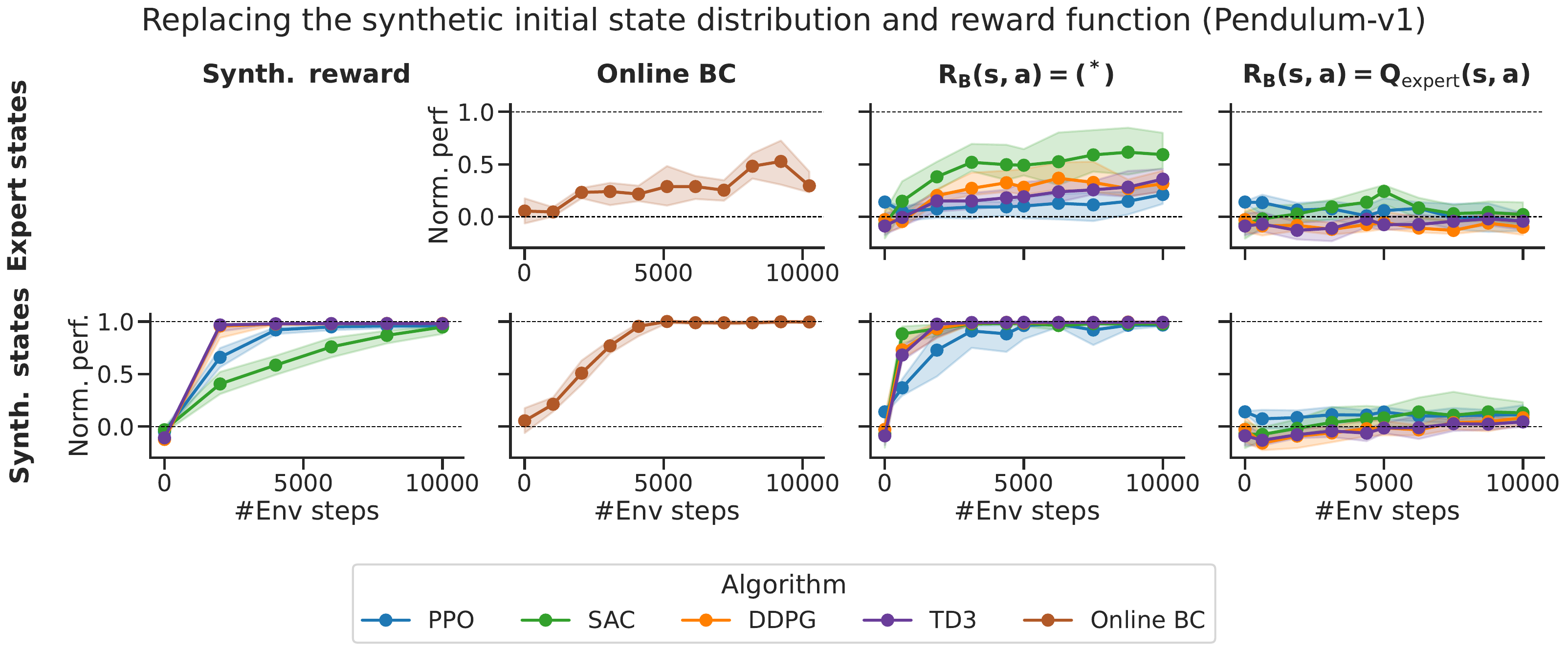}
    \caption{
    Training performance of agents when replacing the synthetic reward function and initial state distribution of an SCB for Pendulum-v1.
    The performance is normalized as $(R - R_\text{expert}) / (R - R_\text{random})$, where $R$, $R_\text{random}$ and $R_\text{expert}$ are the episodic return of the current, a random policy, and an expert policy, respectively. 
    Experts were chosen to be SAC agents trained on the evaluation environment directly.
    Their final returns are shown in table~\ref{tab:sac_expert_cc}.
    We report the IQM of episodic return with 95\% confidence intervals using 20 independent training runs with 50 evaluation rollouts each.
    \textbf{Online BC:} RL training in online behavioral cloning setup, where we take steps towards minimizing $\KL[\pi||\pi_\text{expert}]$ on batches of observations. 
    $\bm{R_B(s, a) = (^*)}$: Training to minimize the error towards the expert action. 
    The reward is computed as $-||a - a^*||$ for continuous, and $\mathbbm{1}[a = a^*]$ for discrete environments, where $\mathbbm{1}$ is the indicator function and $a^*$ is the action chosen by the expert agent.
    $\bm{R_B(s, a) = Q_\text{expert}(s, a)}$: Training in an environment where the reward has been replaced by an expert Q-function, similar to the construction in the proof of lemma~\ref{lemma}.
    }
    \label{fig:replacing_components}
\end{figure}

If $Q^*_B(s, a) = R_B(s, a)$ holds for any CB, can we not construct an optimal CB by setting $R_B(s, a) = Q^*_\text{expert}(s, a)$?
Theoretically, training in such a CB should yield a policy with $Q^*_B(s, a) = Q^*_\text{expert}(s, a)$ (see appendix~\ref{sec:lemma_proof}).
Motivated by this insight, we investigate ablations in which we replace the synthetic reward function and initial state distribution.
While most classic control environments are not strongly affected by such ablations, this is not the case for Pendulum-v1, as shown in figure~\ref{fig:replacing_components}.

\begin{figure}[H]
    \centering
    \includegraphics[width=\textwidth]{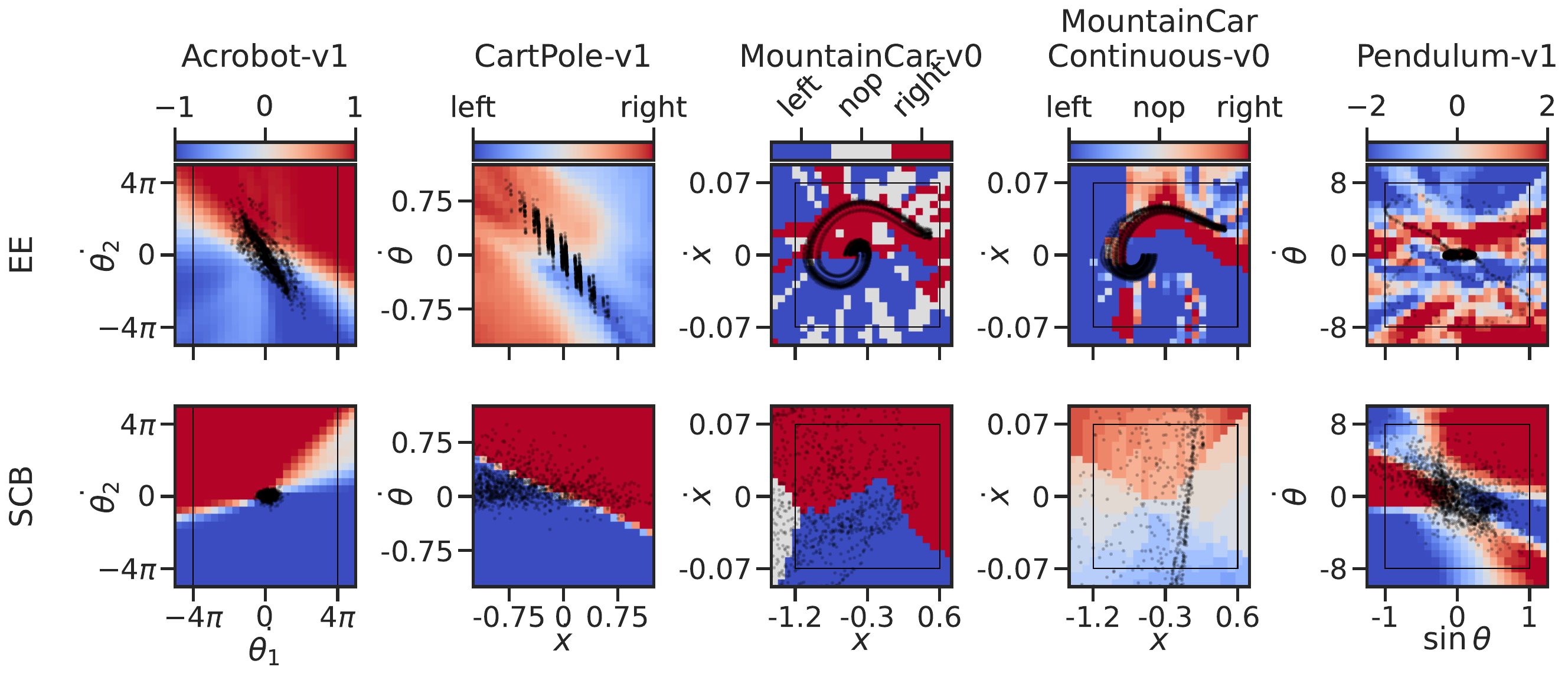}
    \caption{State distributions on the EE and SCB. 
    2000 samples are shown each.
    The state distributions on the EE are generated by an expert SAC policy, for more details see table~\ref{tab:sac_expert_cc}.
    The background indicates $\argmax_a Q_\text{expert}(s, a)$ for the EE (top row), and $\argmax_a R_\text{SCB}(s, a)$ for the SCB (bottom row).
    We observe that the synthetic initial state distribution samples a significantly larger part of the state space than what is visited by the expert.
    Additionally, the synthetic reward functions are smoother for several of the environments.
    We hypothesize that this is because it has to be consistent across all sampled synthetic states, while the Q-function of an expert policy only has to be accurate on its own state distribution.
    }
    \label{fig:state_distributions}
\end{figure}

Firstly, we notice that the synthetic state distribution is needed to train well-performing agents within 10,000 time steps, which is the default for SCBs (bottom row).
We hypothesize that this is because the synthetic initial state distribution samples a large part of the state space, such that agents generalize quickly.
In contrast, the expert state distributions are much less diverse, closely following a small set of trajectories.
Corresponding visualizations are shown in figure~\ref{fig:state_distributions}.
Replacing the synthetic reward by the value function of an expert agent does not work well (column 4).
A possible explanation for this could be that in Pendulum-v1, the difference in Q-values for close actions is small when compared to its absolute magnitude.
Finally, using online behavioral cloning (column 2) or an ``action-supervised'' reward (column 3) in combination with the meta-learned synthetic initial state distribution leads to performance similar to RL training.

\begin{table}[H]
\centering
\begin{tabular}{|l|ccc|}
\hline
Environment & IQM Score & Lower CI & Upper CI \\
\hline
Pendulum-v1 & -137.4 & -151.5 & -127.5 \\
Acrobot-v1 & -76.7 & -78.9 & -74.7 \\
CartPole-v1 & 500.0 & 500.0 & 500.0 \\
ContinuousMountainCar-v0 & 94.9 & 94.8 & 95.0 \\
MountainCar-v0 & -117.7 & -119.2 & -115.3 \\
\hline
\end{tabular}
\caption{IQM and 95\% confidence intervals of the episodic returns achieved by an SAC expert on classic control environments. Aggregated over 200 independent evaluation runs. The agent is near-optimal in all cases.}
\label{tab:sac_expert_cc}
\end{table}

\subsection{Ablations of the Meta-Evolution Algorithm}
\label{sec:meta_evo_ablations}

\begin{figure}[H]
\centering
\includegraphics[width=\textwidth]{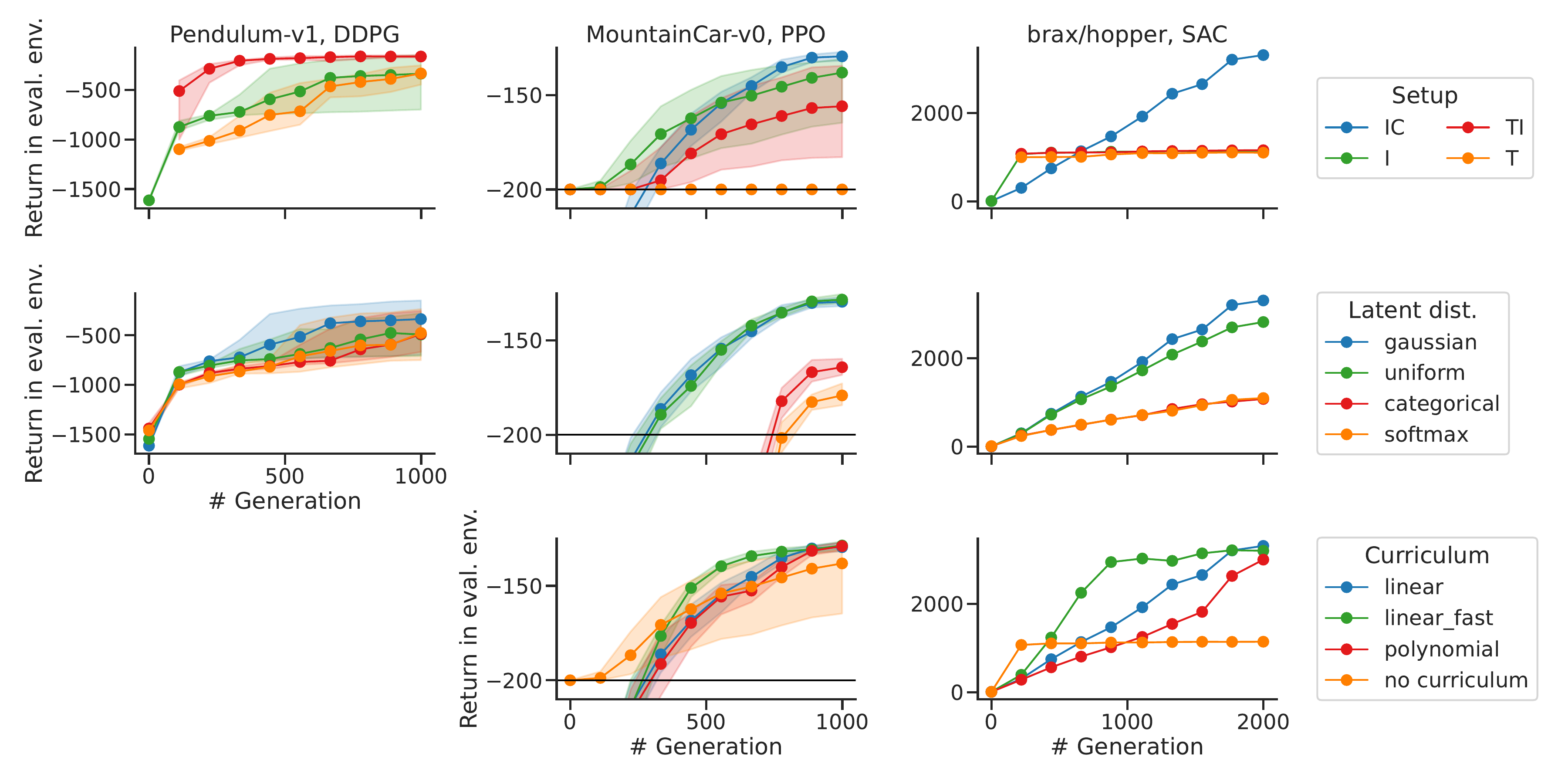}
\caption{
    Ablation study evaluating meta-evolution ingredients. 
    \textbf{Top}. We compare the impact of parameterizing the initial state distribution (I), transition function (T), and the evaluation episode length curriculum (C). 
    All three contributions lead to robust and scalable meta-discovery. 
    \textbf{Middle}. Continuous latent distributions for the initial state distribution perform better than categorical ones. 
    \textbf{Bottom}. The meta-training setup is robust to the exact choice of evaluation episode length curriculum. 
    Inner quantile means for meta-training runs have been averaged over 5 seeds for Pendulum-v1 and 20 seeds for MountainCar-v0, with indicated 95\% confidence intervals.
    }
\label{fig:meta_evo_ablations}
\end{figure}

\pagebreak
Figure~\ref{fig:meta_evo_ablations} shows several ablations of our method.
In the first row, we visualize four different meta-training settings, ingredients indicated by the presence of the letter
\begin{description}
    \item[T] for a parameterized transition function
    \item[I] for a parameterized initial state distribution
    \item[C] for the application of an evaluation episode length curriculum
\end{description}

In the plain T setup, we only parameterize the transition function, which is equivalent to the parameterization of \citet{ferreira2022learning}. 
We use our meta-learning software implementation and hyperparameters to optimize SEs with the T parameterization.
The hyperparameters differ from \citet{ferreira2022learning} only in the population size (increased from 16 to 64-256, depending on the EE) and the number of evaluation seeds (increased from 10 to 64).
Both changes are generally favorable to the performance.
The plain T setup is consistently beaten by our extensions.
On MountainCar-v0, it is not able to discover an environment in which the agent reaches the goal, achieving a mean return of -200 on all evaluation seeds of all meta-training runs.
It is well known that even state-of-the-art RL algorithms such as PPO struggle with solving MountainCar, due to the sparse reward of reaching the flag, which is very improbable to achieve through random exploration.
Having to learn the transition function before training an agent adds another layer of complexity, which makes solving MountainCar in the T setup infeasible.

Introducing a parameterized initial state distribution in TI circumvents this problem, as the environment can learn a distribution of relevant observations directly, without having to reach them via repeated application of the transition function.
This increases the performance on almost all classic control environments, including Pendulum-v1.
We noticed that in T and TI, nearly every generation had members with fitness values of nan.
This leads to missing data points in figure~\ref{fig:meta_evo_ablations}, where even the population mean was affected in Pendulum-v1.
This is because, for long episodes, the recurrent forward pass of synthetic states through the transition neural network can lead to exploding values, which eventually overflow.

This problem can be addressed by limiting the maximum episode length.
Since most episodes are already extremely short in the T and TI setup (typically under 10 time steps) we set the maximum episode length to 1, effectively reducing the synthetic MDP to an SCB task without transition dynamics, leading to the plain I setup. 
We find that this does not reduce the performance in any environment, except for Pendulum-v1, where some meta-training runs converge to a lower value.
Still, the best runs of TI and I have comparable performance.

A curriculum like in IC is needed to achieve competitive results in the Brax environments.
Similar curricula can be introduced to classic control environments.
For example, decreasing the evaluation length from 1000 to 200 while meta-training an environment for MountainCar improves meta-training stability and performance.
The applicability of curricula is specific to each environment.
For Pendulum-v1, it is unclear what kind of curriculum to apply, so we omit it, leaving the bottom left part of figure~\ref{fig:meta_evo_ablations} empty.

Our setup includes two main hyperparameters: the latent distribution from which the initial states are generated and the curriculum.
The second row of figure~\ref{fig:meta_evo_ablations} shows meta-training curves for different latent distributions.
We test four different latent distributions: a standard Gaussian, a uniform distribution in the interval $[0, 1)$, a categorical distribution with equal probabilities, and a categorical distribution with probabilities generated by applying the softmax function to $[1, 2, \dots, n]$, where $n$ is the dimensionality of the latent vector.
When using categorical latent distributions, the initial state distribution becomes a categorical one as well and can be thought of as sampling from a set of meta-learned observations.
Overall, the Gaussian and uniform distributions achieve a similar performance, outperforming the categorical distributions.
This is likely because they can densely sample a manifold of the state space.
The third row of figure~\ref{fig:meta_evo_ablations} shows meta-training curves for different curricula, showing that meta-training is robust to the choice of curriculum.
\newpage

\section{Full Training Results on Brax}
\label{sec:training_speed_comparison_brax}
\begin{figure}[H]
    \centering
    \includegraphics[width=0.98\textwidth]{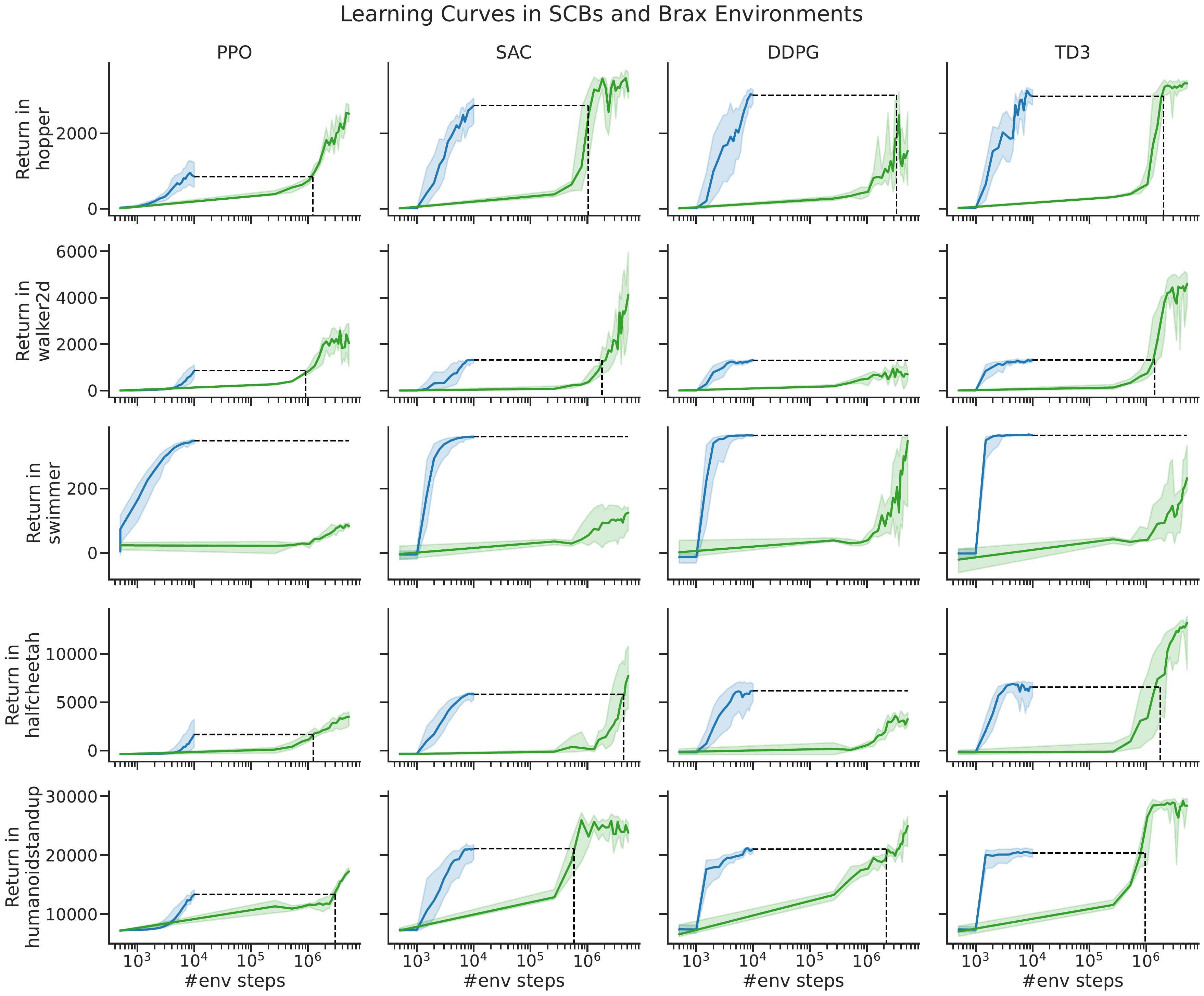}
    \caption{
    Training curves for SCBs and their corresponding Brax environments. Lines show IQMs over 20 runs for the SCB and 5 runs for the EE.
    Shaded areas are 95\% confidence intervals.
    Each of the trained agents is evaluated on 50 episode rollouts.
    EE expert hyperparameters are shown in appendix~\ref{sec:brax_expert_hyperparameters}. 
    Dashed lines are used to indicate the number of steps needed in the EE to match the final performance in the SCB.
    }
    \label{fig:training_scb_real_accumulated}
\end{figure}
\begin{figure}[H]
    \centering
    \includegraphics[width=0.5\textwidth]{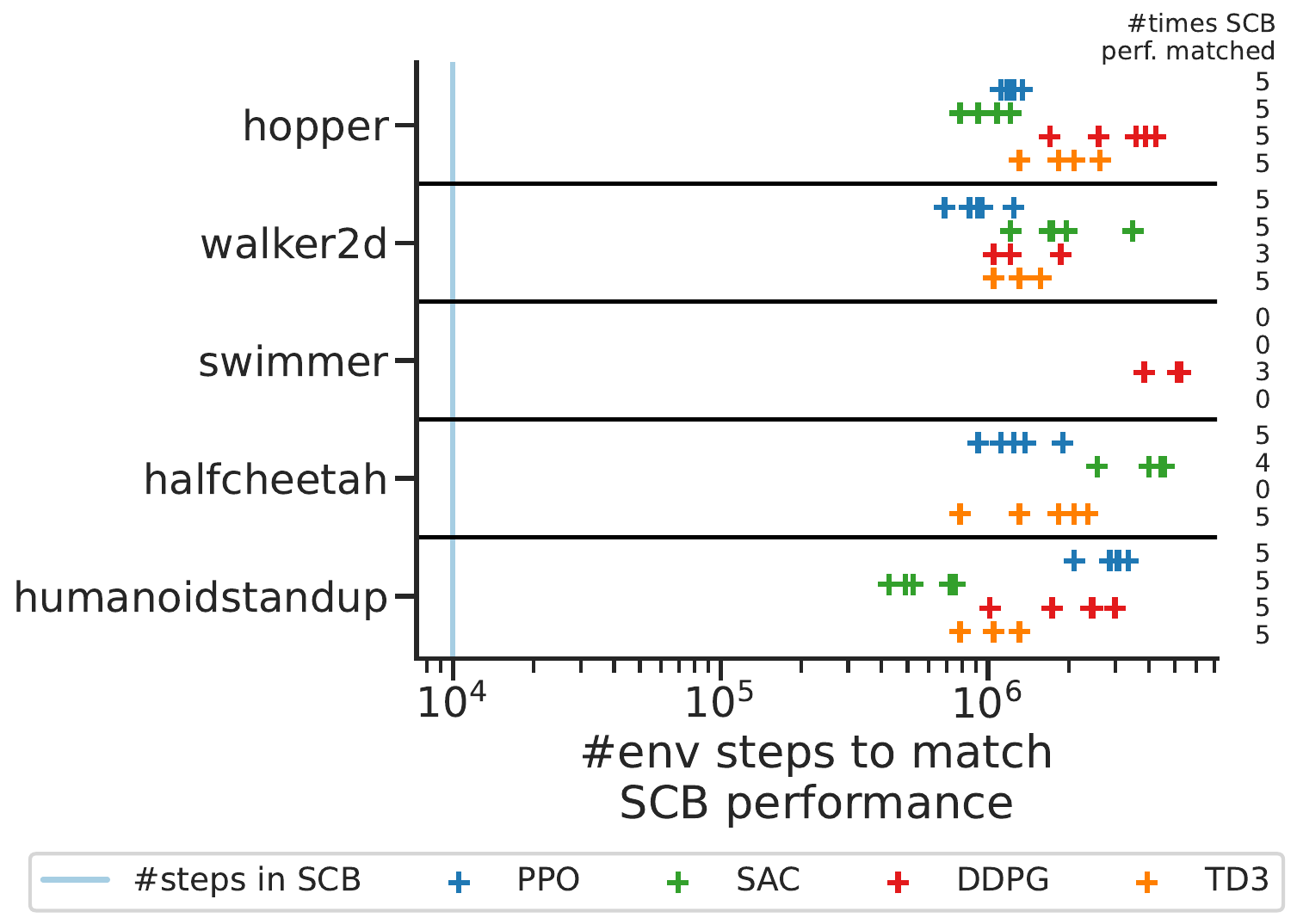}
    \caption{Training in the SCB is two orders of magnitude faster than training in the EE.
    For three algorithms, we train 20 independent agents in an SCB (10k steps) and record the IQM final performance as a baseline. 
    Afterward, we train 5 independent agents on the evaluation environments (5000k steps) using tuned hyperparameters (see appendix~\ref{sec:brax_expert_hyperparameters}).
    For each agent, we record the number of environment steps until it reaches the baseline. 
    We denote the number of times the EE agents match the SCB performance next to each row.}
    \label{fig:match_performance}
\end{figure}
\begin{table}[H]
\scriptsize
    \centering
    \begin{tabular}{|l|c|c|c|c|c|c|c|c|}
    \hline
 & \multicolumn{2}{c|}{PPO} & \multicolumn{2}{c|}{SAC} & \multicolumn{2}{c|}{DDPG} & \multicolumn{2}{c|}{TD3} \\ \cline{2-9} 
Environment & EE  & SCB  & EE  & SCB & EE & SCB & EE & SCB \\
\hline
hopper          & \textbf{2521.9}  & 853.5   & \textbf{3119.4}  & 2738.8  & 1536.0  & \textbf{3012.4}  & \textbf{3325.8}  & 2985.3  \\
walker2d        & \textbf{2039.6}  & 858.3   & \textbf{4140.1}  & 1323.1  & 698.3   & \textbf{1304.3}  & \textbf{4605.8}  & 1321.8  \\
swimmer         & 83.6    & \textbf{348.5}   & 124.8   & \textbf{361.6}   & 348.5   & \textbf{365.1}   & 232.2   & \textbf{365.4}   \\
halfcheetah     & \textbf{3487.1}  & 1657.4  & \textbf{7735.5}  & 5810.4  & 3263.3  & \textbf{6162.4}  & \textbf{13213.5} & 6555.8  \\
humanoidstandup & \textbf{17243.5} & 13356.1 & \textbf{23808.1} & 21105.2 & \textbf{24944.8} & 21039.0 & \textbf{28376.2} & 20372.0 \\
    \hline
    \end{tabular}
    \caption{Final training performance of agents trained in the SCB and EE. We indicate the IQM of the episodic return for 20 independent training runs for the SCB and 5 runs for the EE. The final evaluation is done using 50 episodes.}
    \label{tab:brax_retsults_table}
\end{table}

\section{Hyperparameters}
\label{sec:hyperparameters}
We manually set the outer loop hyperparameters informed by exploratory experiments, since the experiments are very computationally expensive. 
The hyperparameters of the inner loop algorithms were set arbitrarily to a large range of reasonable values. 

We ran all our experiments on four Nvidia A100 80GB GPUs and one Intel Xeon 4215R CPU.
The time to complete one meta-training run depends on the environment, ranging from less than an hour (most classic control environments) to 24 hours (Brax environments, Pendulum-v1).

\subsection{Outer loop hyperparameters}
\label{sec:hyper_outer}

\begin{table}[H]
    \centering
    \scriptsize
    \begin{tabular}{|l|c|c|c|}
        \hline
         & Classic Control & Pendulum & Brax \\
        \hline
        Init. SNES $\sigma$ & 0.05 & 0.05 & 0.05 \\
        Population size & 128 & 64 & 256 \\
        num. rollouts & 1 & 8 & 1\\
        num. eval. seeds & 50 & 50 & 16\\
        \begin{tabular}[x]{@{}c@{}}num. eval. seeds \\for population mean\end{tabular} & 64 & 64 & 64 \\
        multi algo. mode & all & all & sequential \\
        \hline
    \end{tabular}
    \caption{Hyperparameters for meta-training. \textit{multi algo. mode} refers to the way the RL algorithms are chosen in the inner loop. ``All'' means executing all available ones sequentially, and taking the mean of their returns as the fitness. ``Sequential'' means using algorithm $i$ where $i = \text{gen} \mod |A|$.
    }
    \label{tab:hyper_meta}
\end{table}

\begin{table}[H]
    \centering
    \scriptsize
    \begin{tabular}{|l|c|}
        \hline
         & num. generations \\
        \hline
        Acrobot & 300 \\
        CartPole & 300 \\
        MountainCar & 1000 \\
        ContinuousMountainCar & 300 \\
        Pendulum & 1000 \\
        Inverted Pendulum & 300 \\
        Inverted Double Pendulum & 300 \\
        Reacher & 30 \\
        Pusher & 100 \\
        Hopper & 2000 \\
        Walker2D & 2000 \\
        Swimmer & 2000 \\
        Halfcheetah & 2000 \\
        Ant & 2000 \\
        \hline
    \end{tabular}
    \caption{Number of generations for each environment}
    \label{tab:hyper_num_gens}
\end{table}

\begin{table}[H]
    \centering
    \scriptsize
    \begin{tabular}{|l|c|c|c|}
        \hline
         & MountainCar & Brax \\
        \hline
        type & linear & linear \\
        init. eval. length & 1000 & 100 \\
        final eval. length & 200 & 1000 \\
        begin transition & 200 & 200 \\
        num. transitions steps & 600 & 1600\\
        \hline
    \end{tabular}
    \caption{Hyperparameters for evaluation length curricula.}
    \label{tab:hyper_curriculum}
\end{table}

\begin{table}[H]
    \centering
    \scriptsize
    \begin{tabular}{|l|c|}
        \hline
         & all environments \\
        \hline
        network arch. & (32, ) MLP \\
        activation & tanh \\
        latent dist. & $\mathcal{N}(0,I_n)$ \\
        latent size & (dim. of eval. envs. obs. space) \\
        \hline
    \end{tabular}
    \caption{Hyperparameters for the synthetic environment}
    \label{tab:hyper_se}
\end{table}

\subsection{Inner loop hyperparameters}
\label{sec:hyper_inner}
\begin{table}[H]
    \centering
    \scriptsize
    \begin{tabular}{|l|c|c|}
        \hline
         & Classic Control & brax\\
        \hline
        network arch. & (64, 64) MLP & (64, 64) MLP \\
        activation & tanh (ReLU for Pendulum) & tanh \\
        num. envs & 5 & 5 \\
        num. steps & 100 & 100 \\
        num. epochs & 10 & 10 \\
        num. minibatches & 10 & 10  \\
        time steps & $10^4$ & $10^4$ \\
        max. grad. norm & 10 & 10\\
        learning rate & \begin{tabular}[x]{@{}c@{}}$\{0.01, \underline{0.005}, 0.001, 0.0005, 0.0001\}$\\(without 0.01 for continuous environments)\end{tabular} & 0.005\\
        discount & $\{1.0, \underline{0.99}, 0.95, 0.9, 0.8\}$ & 0.99\\
        $\lambda$ for GAE & $\{1.0, \underline{0.95}, 0.9, 0.8, 0.5\}$ & 0.95\\
        clipping $\epsilon$ & $\{0.1, \underline{0.2}, 0.3, 0.4, 0.5\}$ & 0.2\\
        entropy coef. & $\{0.0, \underline{0.01}, 0.05, 0.1, 0.5\}$ & 0.01\\
        value function coef. & $\{0.0, \underline{0.5}, 1.0, 1.5, 2.0\}$ & 0.5\\
        \hline
    \end{tabular}
    \caption{Hyperparameters for PPO in the inner loop. Underlined values are used in runs with a fixed configuration.}
    \label{tab:hyper_se_ppo}
\end{table}

\begin{table}[H]
    \centering
    \scriptsize
    \begin{tabular}{|l|c|c|}
        \hline
         & Classic Control & brax\\
        \hline
        network arch. & (64, 64) MLP & (64, 64) MLP \\
        activation & tanh (ReLU for Pendulum) & tanh \\
        num. envs & 5 & 1 \\
        buffer size & 2000 & 5000 \\
        prefill buffer & 1000 & 1000 \\
        batch size & 256 & 250 \\
        grad. steps & 2 & 1 \\
        time steps & $10^4$ & $10^4$ \\
        learning rate & $\{0.01, \underline{0.005}, 0.001, 0.0005, 0.0001\}$ & 0.005 \\
        discount & $\{1.0, \underline{0.99}, 0.95, 0.9, 0.8\}$ & 0.99 \\
        Polyak $\tau$ & $\{0.99, \underline{0.95}, 0.9, 0.7, 0.8\}$ & 0.95 \\
        target entropy ratio & $\{0.1, 0.3, 0.5, \underline{0.7}, 0.9\}$ & n.a. \\
        \hline
    \end{tabular}
    \caption{Hyperparameters for SAC in the inner loop. Underlined values are used in runs with a fixed configuration.}
    \label{tab:hyper_se_sac}
\end{table}

\begin{table}[H]
    \centering
    \scriptsize
    \begin{tabular}{|l|c|}
        \hline
         & Classic Control (discrete) \\
        \hline
        network arch. & (64, 64) MLP \\
        activation & tanh \\
        num. envs & 10 \\
        buffer size & 2000 \\
        prefill buffer & 1000 \\
        batch size & 100 \\
        grad. steps & 1 \\
        time steps & $10^4$ \\
        target update freq. & 50 \\
        max. grad. norm & 10 \\
        $\epsilon$ start & 1 \\
        $\epsilon$ decay fraction & 0.5 \\
        $\epsilon$ end & $\{0.01, 0.05, 0.1, 0.2\}$ \\
        learning rate & $\{0.01, \underline{0.005}, 0.001, 0.0005, 0.0001\}$ \\
        discount & $\{1.0, \underline{0.99}, 0.95, 0.9, 0.8\}$ \\
        Double DQN & $\{\underline{\text{yes}}, \text{no}\}$ \\
        \hline
    \end{tabular}
    \caption{Hyperparameters for DQN in the inner loop. Underlined values are used in runs with a fixed configuration.}
    \label{tab:hyper_se_dqn}
\end{table}

\begin{table}[H]
    \centering
    \scriptsize
    \begin{tabular}{|l|c|c|}
        \hline
         & Classic Control (continuous) & brax\\
        \hline
        network arch. & (64, 64) MLP & (64, 64) MLP \\
        activation & tanh (ReLU for Pendulum) & tanh \\
        num. envs & 1 & 1 \\
        buffer size & 2000 & 5000 \\
        prefill buffer & 1000 & 1000 \\
        batch size & 100 & 100 \\
        grad. steps & 1 & 1 \\
        time steps & $10^4$ & $10^4$ \\
        max. grad. norm & 10 & 10\\
        learning rate & $\{0.01, \underline{0.005}, 0.001, 0.0005, 0.0001\}$ & 0.005 \\
        discount & $\{1.0, \underline{0.99}, 0.95, 0.9, 0.8\}$ & 0.99 \\
        Polyak $\tau$ & $\{0.99, \underline{0.95}, 0.9, 0.7, 0.8\}$ & 0.95 \\
        expl. noise & $\{0.1, \underline{0.2}, 0.3, 0.5, 0.7, 0.9\}$ & 0.2 \\
        \hline
    \end{tabular}
    \caption{Hyperparameters for DDPG in the inner loop. Underlined values are used in runs with a fixed configuration.}
    \label{tab:hyper_se_ddpg}
\end{table}

\begin{table}[H]
    \centering
    \scriptsize
    \begin{tabular}{|l|c|c|}
        \hline
         & Classic Control (continuous) & brax\\
        \hline
        network arch. & (64, 64) MLP & (64, 64) MLP \\
        activation & tanh (ReLU for Pendulum) & tanh \\
        num. envs & 1 & 1 \\
        buffer size & 2000 & 5000 \\
        prefill buffer & 1000 & 1000 \\
        batch size & 100 & 100 \\
        grad. steps & 1 & 1 \\
        time steps & $10^4$ & $10^4$ \\
        max. grad. norm & 10 & 10\\
        learning rate & $\{0.01, \underline{0.005}, 0.001, 0.0005, 0.0001\}$ & 0.005 \\
        discount & $\{1.0, \underline{0.99}, 0.95, 0.9, 0.8\}$ & 0.99 \\
        Polyak $\tau$ & $\{0.99, \underline{0.95}, 0.9, 0.7, 0.8\}$ & 0.95 \\
        expl. noise & $\{0.1, \underline{0.2}, 0.3, 0.5, 0.7, 0.9\}$ & 0.2 \\
        target noise & $\{0.1, \underline{0.2}, 0.3, 0.5, 0.7, 0.9\}$ & 0.2 \\
        target noise clip & $\{0.1, 0.4, \underline{0.5}, 0.7, 1.0, 1.3\}$ & 0.5 \\
        \hline
    \end{tabular}
    \caption{Hyperparameters for TD3 in the inner loop. Underlined values are used in runs with a fixed configuration.}
    \label{tab:hyper_se_td3}
\end{table}

\subsection{Brax Expert Hyperparameters}
\label{sec:brax_expert_hyperparameters}
We train several expert agents in Brax environments.
Our goal for the expert agents is not to match the state-of-the-art results, but instead, to get a baseline that represents good performance.
We therefore fit hyperparameters using a budget of 20 runs using random search, using a fixed number of environment steps, buffer size, the number of vectorized environments, and others.

\begin{table}[H]
\centering
    \scriptsize
\begin{tabular}{|l|c|c|c|c|c|}
\hline
Parameter & Hopper & Walker2d & Swimmer & Halfcheetah & Humanoidstandup \\
\hline
learning rate & $2.5\cdot 10^{-4}$ & $1.7 \cdot 10^{-3}$ & $2.8 \cdot 10^{-3}$ & $1.7 \cdot 10^{-3}$ & $1.7 \cdot 10^{-3}$\\
num. envs. & 32 & 32 & 128 & 32 & 32 \\
num. steps & 32 & 32 & 64 & 32 & 32 \\
num. epochs & 9 & 2 & 7 & 2 & 2 \\
num. minibatches & 2 & 2 & 8 & 2 & 2 \\
discount & 0.995 & 0.99 & 0.98 & 0.99 & 0.99 \\
$\lambda$ for GAE & 0.95 & 0.8 & 0.99 & 0.8 & 0.8 \\
max. grad. norm & 1 & 5 & 0.5 & 5 & 5 \\
\hline
network arch. & \multicolumn{5}{c|}{(64, 64) MLP} \\
activation  & \multicolumn{5}{c|}{tanh} \\
time steps & \multicolumn{5}{c|}{$5 \cdot 2^{20}$} \\
clip\_eps & \multicolumn{5}{c|}{0.2} \\
entropy coef. & \multicolumn{5}{c|}{0.01} \\
value function coef. & \multicolumn{5}{c|}{0.5} \\
norm. obs. & \multicolumn{5}{c|}{true} \\
\hline
\end{tabular}
\caption{PPO expert hyperparamters. Tuned over 20 runs of random search.}
\label{tab:ppo_expert_brax}
\end{table}

\begin{table}[H]
\centering
    \scriptsize
\begin{tabular}{|l|c|c|c|c|c|}
\hline
Parameter & Hopper & Walker2d & Swimmer & Halfcheetah & Humanoidstandup \\
\hline
learning rate & $5.7\cdot 10^{-4}$ & $2\cdot 10^{-4}$ & $5.7\cdot 10^{-4}$ & $2 \cdot 10^{-4}$ & $2 \cdot 10^{-4}$\\
batch size & 128 & 256 & 128 & 512 & 256 \\
discount & 0.995 & 0.99 & 0.995 & 0.98 & 0.99 \\
Polyak $\tau$ & 0.95 & 0.99 & 0.95 & 0.995 & 0.99 \\
\hline
network arch. & \multicolumn{5}{c|}{(64, 64) MLP} \\
activation & \multicolumn{5}{c|}{tanh} \\
num. envs. & \multicolumn{5}{c|}{128} \\
buffer size & \multicolumn{5}{c|}{$2^{20}$} \\
prefill buffer & \multicolumn{5}{c|}{$2^{13}$} \\
grad. steps & \multicolumn{5}{c|}{128} \\
time steps & \multicolumn{5}{c|}{$5 \cdot 2^{20}$} \\
norm. obs. & \multicolumn{5}{c|}{true} \\
\hline
\end{tabular}
\caption{SAC expert hyperparameters. Tuned over 20 runs of random search.}
\label{tab:sac_expert_brax}
\end{table}

\begin{table}[H]
\centering
    \scriptsize
\begin{tabular}{|l|c|c|c|c|c|}
\hline
Parameter & Hopper & Walker2d & Swimmer & Halfcheetah & Humanoidstandup \\
\hline
learning rate & $1.4 \cdot 10^{-4}$ & $1.4 \cdot 10^{-4}$ & $4.6 \cdot 10^{-4}$ & $2 \cdot 10^{-4}$ & $2 \cdot 10^{-4}$\\
discount & 0.995 & 0.995 & 0.995 & 0.95 & 0.95 \\
Polyak $\tau$ & 0.98 & 0.98 & 0.995 & 0.99 & 0.99\\
batch size & 512 & 512 & 128 & 512 & 512 \\
max. grad. norm & 1 & 1 & 0.1 & 0.1 & 0.1 \\
expl. noise & 0.2 & 0.9 & 1 & 0.5 & 0.5 \\
norm. obs. & true & true & false & true & true \\
\hline
network arch. & \multicolumn{5}{c|}{(64, 64) MLP} \\
activation & \multicolumn{5}{c|}{tanh} \\
num. envs. & \multicolumn{5}{c|}{128} \\
buffer size & \multicolumn{5}{c|}{$2^{20}$} \\
prefill buffer & \multicolumn{5}{c|}{$2^{13}$} \\
gradient steps & \multicolumn{5}{c|}{128} \\
time steps & \multicolumn{5}{c|}{$5 \cdot 2^{20}$} \\
\hline
\end{tabular}
\caption{DDPG expert hyperparamters. Tuned over 20 runs of random search.}
\label{tab:ddpg_expert_brax}
\end{table}

\begin{table}[H]
\centering
    \scriptsize
\begin{tabular}{|l|c|c|c|c|c|}
\hline
Parameter & Hopper & Walker2d & Swimmer & Halfcheetah & Humanoidstandup \\
\hline
learning rate     & $1.8 \cdot 10^{-4}$ & $1.8 \cdot 10^{-4}$ & $2.1 \cdot 10^{-4}$ & $1.2 \cdot 10^{-4}$ & $1.5 \cdot 10^{-4}$\\
discount          & 0.995 & 0.995 & 0.995 & 0.99 & 0.99 \\
Polyak $\tau$     & 0.95 & 0.95 & 0.98 & 0.99 & 0.95 \\
batch size        & 256 & 256 & 512 & 512 & 256 \\
max. grad. norm   & 2 & 2 & 0.1 & 5 & 0.2 \\
expl. noise       & 0.5 & 0.5 & 0.9 & 0.3 & 0.8 \\
target noise      & 0.8 & 0.8 & 0.9 & 0.8 & 1.0 \\
target noise clip & 0.5 & 0.5 & 0.9 & 0.6 & 0.9 \\
policy delay      & 3 & 3 & 1 & 1 & 10 \\
\hline
network arch. & \multicolumn{5}{c|}{(64, 64) MLP} \\
activation & \multicolumn{5}{c|}{tanh} \\
num. envs. & \multicolumn{5}{c|}{128} \\
buffer size & \multicolumn{5}{c|}{$2^{20}$} \\
prefill buffer & \multicolumn{5}{c|}{$2^{13}$} \\
gradient steps & \multicolumn{5}{c|}{128} \\
time steps & \multicolumn{5}{c|}{$5 \cdot 2^{20}$} \\
norm. obs. & \multicolumn{5}{c|}{true} \\
\hline
\end{tabular}
\caption{TD3 expert hyperparamters. Tuned over 20 runs of random search.}
\label{tab:td3_expert_brax}
\end{table}



\end{document}